\documentclass[10pt,twocolumn,letterpaper]{article}

\usepackage{cvpr}              

\usepackage{pythonhighlight}
\usepackage{pythontex}
\usepackage{multirow}
\usepackage{amsmath}
\usepackage{pifont}
\usepackage{subcaption}
\usepackage[table]{xcolor}
\usepackage[accsupp]{axessibility}
\newcommand{\pub}[1]{\color{gray}{\tiny{#1}}}

\definecolor{mjjeonc}{RGB}{34,139,34}
 
\definecolor{shchoic}{RGB}{177, 44, 0}

\definecolor{jhleec}{RGB}{0, 185, 0}

\definecolor{cvprblue}{rgb}{0.21,0.49,0.74}
\definecolor{Grey}{HTML}{ededed}
\definecolor{Blue}{HTML}{ecf4f8}
\usepackage[pagebackref,breaklinks,colorlinks,allcolors=cvprblue]{hyperref}
\AtEndPreamble{
    \crefname{section}{Sec.}{Secs.}
    \Crefname{section}{Section}{Sections}
    \Crefname{table}{Table}{Tables}
    \crefname{table}{Tab.}{Tabs.}
    \Crefname{figure}{Figure}{Figures}
}

\def\paperID{43150} 
\def\confName{CVPR}
\def\confYear{2026}

\title{Follow the Saliency:  \\ Supervised Saliency for Retrieval-augmented
Dense Video Captioning
}

\author{
    Seung hee Choi \quad 
    MinJu Jeon \quad 
    Hyunwoo Oh \quad 
    Jihwan Lee \quad 
    Dong-Jin Kim \\
    Hanyang University \\
    {\tt\small \{ermitaju1, mnju5026, komjii, jjt2345, djdjkim\}@hanyang.ac.kr}
}

\begin{document}
\maketitle

\begin{abstract}
Existing retrieval-augmented approaches for Dense Video Captioning (DVC) often fail to achieve accurate temporal segmentation aligned with true event boundaries, as they rely on heuristic strategies that overlook ground truth event boundaries.
The proposed framework, \textbf{STaRC}, overcomes this limitation by supervising frame-level saliency through a highlight detection module. 
Note that the highlight detection module is trained on binary labels derived directly from DVC ground truth annotations without the need for additional annotation.
We also propose to utilize the saliency scores
as a unified temporal signal that drives retrieval via saliency-guided segmentation and informs caption generation through explicit Saliency Prompts injected into the decoder.
By enforcing saliency-constrained segmentation, our method produces temporally coherent segments that align closely with actual event transitions, leading to more accurate retrieval and contextually grounded caption generation. We conduct comprehensive evaluations on the YouCook2 and ViTT benchmarks, where STaRC achieves state-of-the-art performance across most of the metrics. Our code is available at \url{https://github.com/ermitaju1/STaRC}

\end{abstract}

\section{Introduction}
\label{sec:1_intro}

\begin{figure}[t]
    \centering
    \includegraphics[width=0.98\linewidth]{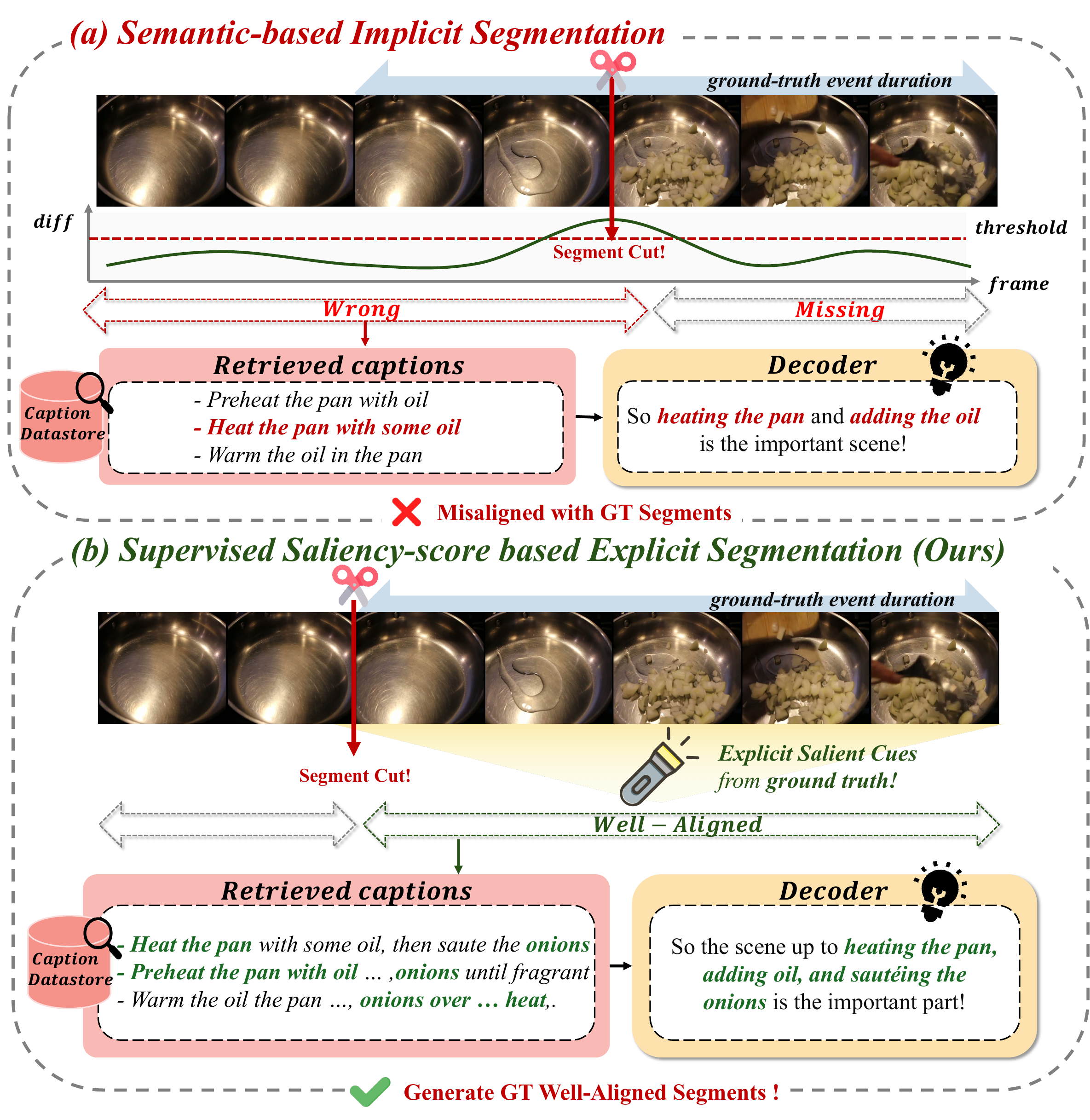}
    \caption{
\textbf{An example of caption retrieval and generation by existing methods} (e.g., Sali4Vid~\cite{jeon2025sali4vid}) and ours. The misaligned video segments in existing methods lead to less relevant retrieval results. Our supervised saliency scores are well-aligned with the ground truth, enabling more accurate retrieval. This alignment also provides the decoder with more precise and contextually appropriate information.}

\label{fig:teaser_fig}
\end{figure}


\begin{figure*}[t]
\begin{center}
\includegraphics[width=.9\textwidth]{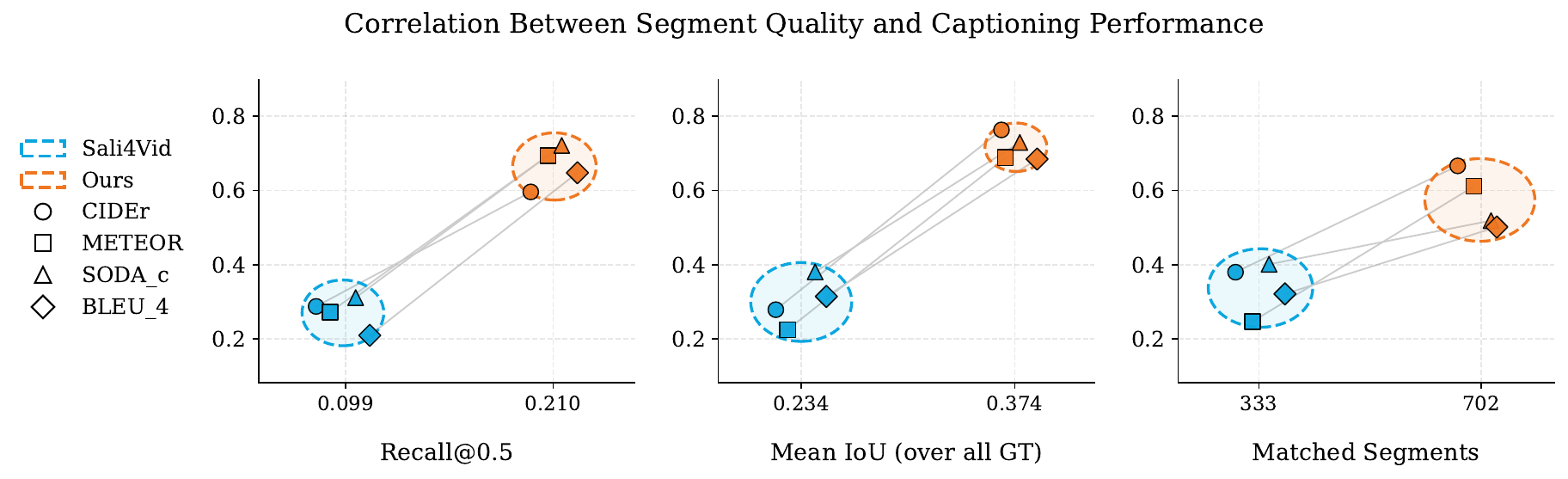}
\caption{\textbf{Correlation analysis between segment-quality metrics and captioning performance}. The x-axis shows three indicators of temporal localization quality - \textit{Recall@0.5}, \textit{Mean IoU}, and \textit{Matched Segments} - which respectively measure the proportion of correctly detected events (IoU $\ge 0.5$), the average overlap with ground truth segments, and the number of predictions aligned with the reference events. As these segment-quality metrics improve from Sali4Vid~\cite{jeon2025sali4vid} to ours, downstream DVC performances (e.g., CIDEr, METEOR) also rise consistently, demonstrating a strong positive correlation between accurate event segmentation and caption generation performance.} 
\label{fig:teaser_graph}
\end{center}
\end{figure*}

Dense Video Captioning (DVC) aims to detect and describe multiple events across long, untrimmed videos~\cite{li2018dense, wei2023dense, duan2018dense, zhou2018dense, dvc, mkhallati2023dense}.
This is fundamentally different from standard video captioning~\cite{gao2017video, chen2017video, wang2018video, seo2022video, zhao2023video}, which produces a single sentence for short, trimmed clips.
Because DVC requires modeling a sequence of semantically rich events, recent studies have shown that visual-only approaches often struggle to capture the necessary event-level structure~\cite{pdvc}. Consequently, retrieval-augmented methods~\cite{cm2, kim2025hicm2, jeon2025sali4vid} have gained attention.
These approaches enhance the decoder's understanding by retrieving relevant captions from external datastores and have achieved strong performance on YouCook2~\cite{youcook2} and ViTT~\cite{vitt}.
Despite their success, retrieval operates on video segments formed by clustering frame-level features, making the quality of segmentation directly affect the overall captioning performance.

As shown in \Cref{fig:teaser_fig} (a), implicit segmentation strategies in prior methods often produce misaligned segments that fail to capture true event boundaries. For instance, uniform sampling in  HiCM$^2$~\cite{kim2025hicm2} and similarity-based clustering in Sali4Vid~\cite{jeon2025sali4vid} frequently group unrelated frames together, leading the model to retrieve less relevant captions. A typical failure case retrieves ``\textit{Preheat 
the pan with oil}'' instead of the correct event ``\textit{heat the pan with some oil, 
then saut\'{e} the onions}.''. Such boundary errors propagate to retrieval, causing the decoder to receive less relevant context.
In contrast, \Cref{fig:teaser_fig}(b) illustrates how improved segment formation can yield boundaries that align more closely with true events. These higher quality segments provide retrieval modules with input that more accurately reflects the underlying action, resulting in more relevant retrieved captions.

This relationship is further validated by our correlation analysis in \Cref{fig:teaser_graph}, which shows that improvements in segment-quality indicators (e.g., Recall@0.5, Mean IoU, Matched Segments) are strongly associated with higher DVC evaluation metric such as CIDEr and METEOR. When segments more closely align with true event boundaries, they supply the decoder with more meaningful retrieval information, thereby consistently enhancing captioning performance. These findings emphasize the need for a framework that improves segment alignment 
with ground truth to better guide the decoder.

To address these limitations, we propose \textbf{\textit{S}}aliency \textbf{\textit{T}}r\textbf{\textit{a}}ining for \textbf{\textit{R}}etrieval and \textbf{\textit{C}}aptioning (\textit{\textbf{STaRC}}), a supervised saliency learning framework for retrieval-augmented DVC. Our key idea is \textbf{\textit{to learn frame-level saliency scores that reflect frame importance from ground truth event annotations}} using a highlight detection module following prior work~\cite{xiao2024bridging, um2025watch}. The labels for this module are derived directly from annotations, requiring no additional supervision. Unlike prior approaches that implicitly use saliency during generation, we treat it as 
an explicit and shared supervisory signal for both retrieval and captioning. We introduce 
a \textbf{\textit{Unified Saliency-Based Design}} that leverages the learned saliency 
scores in two complementary ways.

First, we introduce \textbf{\textit{Saliency-Guided Segmentation and Retrieval (SGSR)}} 
to form event-aligned segments. \textit{SGSR} uses Optimal Transport~\cite{xu2024temporally} 
to cluster frames based on predicted saliency scores. Previous method~\cite{jeon2025sali4vid} 
clusters frames using only semantic similarity without ground truth guidance. In contrast, 
\textit{SGSR} uses saliency scores learned from annotated event boundaries. This produces segments that align with true event boundaries to better separate distinct events. These segments enable the model to retrieve more relevant captions for the decoder. 

Second, we introduce \textbf{\textit{Saliency Prompt (SaliP)}}, which explicitly 
incorporates saliency into the decoder. Previous method~\cite{jeon2025sali4vid} implicitly applies saliency by multiplying 
video features with temporal masks. In contrast, \textit{SaliP} directly injects frame-level saliency scores as grounding prompts~\cite{wang2025visual} into the decoder. This 
explicit guidance enables the decoder to focus on semantically important frames. Through this unified design, \textit{STaRC} 
achieves coherent alignment between retrieval and caption generation.

The main contributions of our framework are as follows:
\begin{itemize}
\item We propose \textbf{\textit{STaRC}}, a supervised saliency learning framework for retrieval-augmented DVC. Our model explicitly learns frame-level saliency scores using a highlight detection module trained with binary labels converted from ground truth event annotations, without any extra annotation cost.

\item We introduce a \textbf{unified saliency-based design} that leverages supervised saliency for both retrieval and captioning. Within this design, \textbf{\textit{SGSR}} forms event-aligned segments that improve caption retrieval, and \textbf{\textit{SaliP}} injects saliency into the decoder as temporal grounding.

\item With this unified approach, \textbf{\textit{STaRC}} achieves state-of-the-art performance on YouCook2 and ViTT, demonstrating the benefit of supervised saliency learning for DVC.
\end{itemize}

\begin{figure*}[t]
\begin{center}
\includegraphics[width=0.98\textwidth]{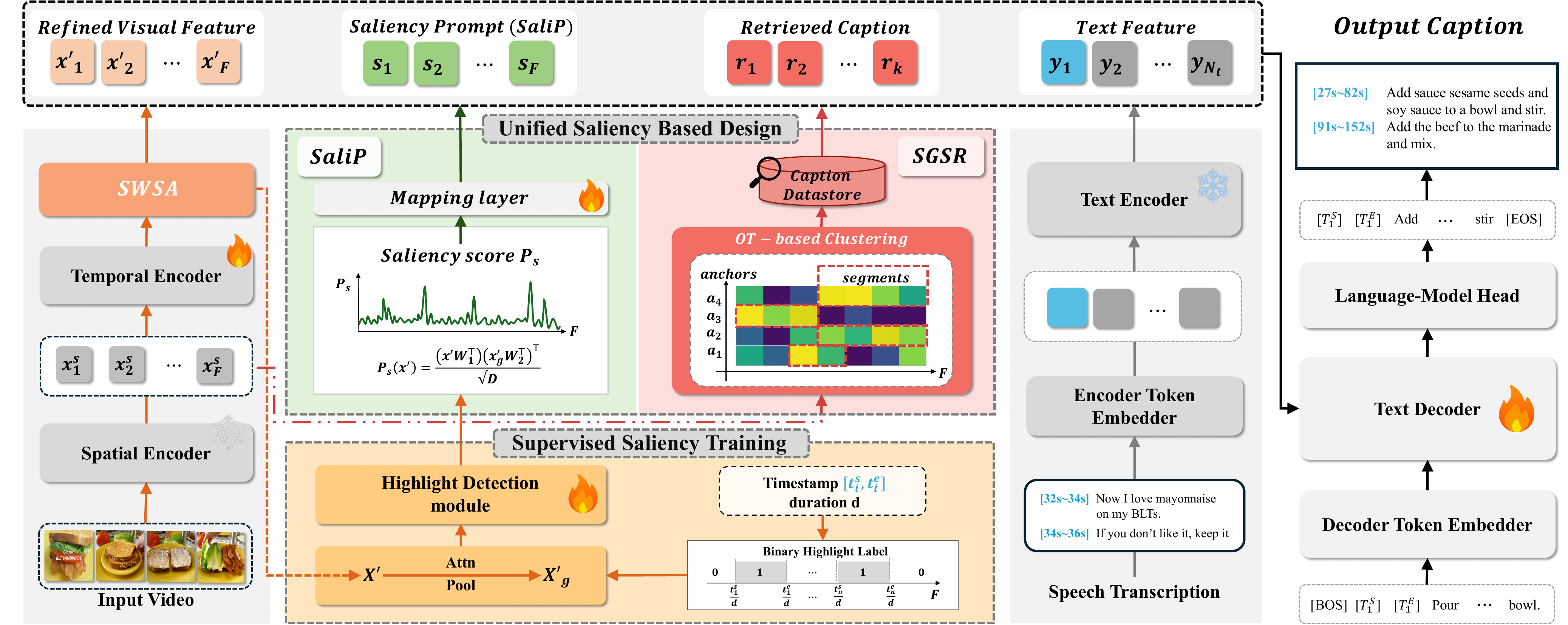}
\end{center}
\caption{\textbf{Overview of our \textit{STaRC} framework for DVC.} A \textit{SWSA} module refines video features, and a highlight detection module is supervised using binary highlight labels derived from existing DVC annotations to predict frame-level saliency scores. The \textit{SGSR} module then performs OT-based clustering guided by these saliency scores to form coherent retrieval segments. In addition, the \textit{SaliP} integrates saliency into the decoder’s attention for saliency-aware caption generation. \textit{STaRC} unifies retrieval and caption generation using saliency signals learned in a supervised manner, ensuring consistent alignment between visual context and semantic description.}

\label{fig:main_figure}
\end{figure*}

\section{Related Works}
\label{sec:related_wokrs}
\subsection{Dense Video Captioning}
Dense Video Captioning (DVC) aims to detect multiple events in long videos and generate natural language descriptions for each~\cite{dvc}. Early two-stage approaches~\cite{better, multi} treated event localization and sentence generation independently, which often caused temporal inconsistencies due to limited cross-stage interaction. End-to-end architectures have been proposed to mitigate these issues. PDVC~\cite{pdvc} formulates DVC as set-level prediction using a DETR-style transformer~\cite{carion2020end}, while Vid2Seq~\cite{yang2023vid2seq} adopts a sequence-to-sequence framework that jointly models localization and captioning. Streaming V2S~\cite{streaming} enables online inference, and DIBS~\cite{dibs} leverages large-scale pseudo-labeled datasets for pretraining. More recent end-to-end methods address structural limitations without relying on large-scale pretraining. E$^2$DVC~\cite{wu2025event} groups frames into pseudo-events using agglomerative clustering to prevent the model from overlooking short or infrequent events. CACMI~\cite{jia2025explicit} also forms pseudo-events through clustering, but uses retrieved captions as queries to align visual and textual features. Both methods show that explicitly modeling event structure improves performance, though their pseudo-events are formed without ground truth guidance, so segment boundaries may not match annotated events. 

Among pretrained approaches, retrieval-based frameworks~\cite{cm2, kim2025hicm2, jeon2025sali4vid} improve visual grounding by retrieving similar captioned segments from external datastores. However, these models remain sensitive to temporal misalignment because they rely on heuristic or fixed segmentation. HiCM$^2$~\cite{kim2025hicm2} constructs fixed-length segments via uniform sampling, and Sali4Vid~\cite{jeon2025sali4vid} uses frame similarity changes to derive boundaries. Both approaches provide coarse temporal cues but lack precise event-level modeling. Recently, Sali4Vid demonstrates that temporally important frames benefit both retrieval and caption generation. However, its saliency is derived heuristically from timestamps rather than learned directly. The potential of supervised saliency learning that unifies retrieval and caption generation remains underexplored in DVC.

\section{Method}
This work presents \textbf{\textit{S}}aliency \textbf{\textit{T}}r\textbf{\textit{a}}ining for \textbf{\textit{R}}etrieval and \textbf{\textit{C}}aptioning (\textit{\textbf{STaRC}}), a framework for supervised saliency learning in Dense Video Captioning (DVC). 
Highlights annotated event boundaries serve as the important temporal regions from which frame-level saliency scores are inferred.

Therefore, frame features are refined using Sliding-Window Self-Attention (\textit{SWSA}), which aggregates local temporal context. The refined features are then passed to the highlight detection module to predict saliency scores.

These scores are integrated through a Unified Saliency-Based Design comprising two components: Saliency-Guided Segmentation and Retrieval (\textit{SGSR}) for ground truth event aligned segment retrieval, and Saliency Prompts (\textit{SaliP}) to guide the decoder toward salient temporal regions during decoding.
The full architecture is shown in \cref{fig:main_figure}.
\subsection{Preliminaries}
\label{sec:preliminaries}

\paragraph{Task Formulation.}

Given an input video consisting of $F$ frames, spatial embeddings $X^s = \{x_n^s\}_{n=1}^F\in \mathbb{R}^{F\times D}$ are extracted using a frozen CLIP ViT-L/14~\cite{radford2021learning}. 
A temporal transformer encoder contextualizes these embeddings into $X = \{x_n\}_{n=1}^F\in \mathbb{R}^{F\times D}$.
The model also receives a transcribed text sequence $Y = \{ y_m \}_{m=1}^{N_t}\in\mathbb{R}^{N_t\times D}$ as input. 
The output is a set of $L$ events $Z = \{ z_q \}_{q=1}^{L}$, where each event $z_q$ consists of its temporal boundaries $(t_q^{\text{start}}, t_q^{\text{end}})$ and its corresponding natural language caption.

\paragraph{Retrieval-augmentation for DVC.}

For retrieval, spatial embeddings $X^{s}$ are used due to their robustness to appearance variations.
The video is divided into $N_{\text{seg}}$ segments, and each segment is represented by an aggregated embedding.
For all segment, the model retrieves the top-$p$ most relevant captions from an external datastore using cosine similarity.
The retrieved captions are encoded with a CLIP text encoder, and their averaged embedding forms a retrieval vector $r_s$ for that segment.
Collectively, these retrieval vectors form $R = \{ r_s \}_{s=1}^{N{\text{seg}}}$, which are provided as auxiliary input to the decoder during caption generation.

\subsection{Saliency Prediction via Highlight Detection}
\label{sec:HD_train}

We define \textit{highlights} as the event boundaries annotated in the DVC dataset. Frame-level \textit{saliency scores} are estimated to measure the semantic importance of each frame within these highlighted intervals. To enable saliency prediction, frame features are first refined using a \textit{SWSA} mechanism. The refined features are then passed to a highlight detection module, which outputs saliency scores.

\paragraph{Feature Refinement.}
\label{sec:SWSA}

The Sliding-Window Self-Attention (\textit{SWSA}) module enhances the temporally encoded frame features before saliency prediction. It applies local self-attention across multiple sliding windows $\{w_1, w_2, w_3\}$ of varying sizes, allowing each frame to incorporate contextual information from neighboring regions at different scales. Unlike standard self-attention, the queries, keys, and values are directly derived from the input features without any linear projections, making the module free of learnable parameters. As the windows slide over the sequence, overlapping outputs are averaged based on their overlap counts. The aggregated features are then normalized and added to the original input $X$ via a residual connection, yielding the refined representation $X'$. The detailed formulation and architecture of \textit{SWSA} are provided in the Appendix.
\paragraph{Supervised Saliency Learning.}
\label{sec:Supervised Saliency Learning}
We use the refined local features $X'$ to obtain frame-level saliency scores in a supervised manner. We adopt a highlight detection module to predict these scores. To capture both fine-grained temporal features and overall video context, the highlight detection module takes both local and global video features as input~\cite{xiao2024bridging, um2025watch}, which predicts the frame-level saliency distribution. A global video feature $X'_{\text{g}} \in \mathbb{R}^{1 \times D}$ is obtained via attention pooling over $X'$, summarizing the overall context of the video. We project $X'$ and $X'_{\text{g}}$ into a shared embedding space through learnable matrices $\mathbf{W}_1$ and $\mathbf{W}_2$. 
The saliency score for each frame is computed as:
\begin{equation}
P_s(x_n') =  \frac{(x_n'\mathbf{W}_1^{\top})({x'_n}_{\text{g}}\mathbf{W}_2^{\top})^{\top}}{\sqrt{D}}.
\label{eq:saliency}
\end{equation}
The resulting $P_s \in \mathbb{R}^{F \times 1}$ represents the frame-level saliency scores, which quantify the relative importance of each frame with respect to the global video context.

\paragraph{Training Objective.}
Since DVC datasets do not contain suitable labels for training a highlight detection module, we convert their annotated event boundaries into binary frame-level highlight labels following the YouTube Highlights protocol~\cite{sun2014ranking}.

Unlike YouTube Highlights, which defines highlights at the clip-level with binary labels, we extend this idea to the frame-level and train the model to predict saliency scores.
Frames within annotated event boundaries are considered highlights and labeled $1$, whereas the others are labeled $0$, producing $\mathbf{H} \in \{0,1\}^{F}$. Because these highlight labels are directly derived from existing DVC annotations, no additional annotation cost is required. Based on these binary labels, we train the highlight detection module using a listwise softmax objective. Given predicted saliency scores ${P}_s$, binary highlight labels $\mathbf{H}$, and a validity mask $\mathbf{M}\in\mathbb{R}^F$, we define the masked softmax probability for each frame $l$ as:
\begin{equation}
p_{l}
=
\frac{\exp({P_s}_l/\tau)\,\mathbf{M}_{l}}
     {\sum_{n=1}^{F} \exp({P_s}_n/\tau)\,\mathbf{M}_{n}}\,,
\label{eq:softmax_prob_single}
\end{equation}
and the video-level saliency loss as:
\begin{equation}
\mathcal{L}_{\text{saliency}}
= -\frac{1}{\sum_{l=1}^{F}\mathbf{H}_{l}\mathbf{M}_{l}}
   \sum_{l=1}^{F}\mathbf{H}_{l}\mathbf{M}_{l}\,\log p_{l}.
\label{eq:video_loss_single}
\end{equation}
As shown in~\Cref{eq:softmax_prob_single}, the normalized softmax distribution enables all frames to compete for probability mass within a video. Building upon~\Cref{eq:video_loss_single}, the objective encourages annotated highlights to receive higher probabilities while implicitly down-weighting unannotated frames.
Note that the ground truth highlight labels $\mathbf{H}$ and $\mathbf{M}$ can be obtained from annotations without any extra labeling cost.

\subsection{Unified Saliency-Based Design}
\label{sec:Unified Saliency-Based Design}
Building on the saliency scores $P_s$ obtained from~\Cref{sec:HD_train}, we introduce a \textit{Unified Saliency-Based Design} that exploits $P_s$ as a shared temporal signal. \textit{STaRC} ensures consistent temporal grounding by using the same saliency scores across both retrieval and caption generation: \textit{SGSR} forms segments based on saliency, while \textit{SaliP} guides the decoder to attend to the same 
salient regions.

\paragraph{Saliency-Guided Segmentation and Retrieval.}
\label{sec:SGSR}
\textit{SGSR} generates video segments that closely align with ground truth event transitions via saliency-guided clustering. These top-ranked segments are then used to retrieve semantically relevant captions from a datastore, providing the decoder with comprehensive contextual information. 

To effectively cluster frames while incorporating saliency information, we 
adopt an Optimal Transport (OT)-based clustering approach inspired by 
ASOT~\cite{xu2024temporally}. We define $K$ learnable anchors 
$A = \{a_j\}_{j=1}^K \in \mathbb{R}^{K \times D}$ as semantic prototypes, 
where each frame is assigned to the anchor with the highest cosine similarity. 
We normalize saliency scores $P_s$ with a sigmoid function to obtain the 
saliency prior $p_s\in[0,1]^F$.

We integrate saliency in two ways. 
{First}, unlike ASOT, which relaxes the anchor side marginal, we apply unbalanced OT on the frame side by using $p_s$ as a 
soft constraint on the frame marginal distribution. Since not all frames contain events in DVC, this allows salient frames to receive higher transport 
mass while keeping anchor marginals balanced. Let $\mathbf{T} \in \mathbb{R}^{F \times K}$ denote the 
transport plan. We solve:
\begin{equation}\label{eq:fugw_prob}
\min_{\mathbf{T}\in\mathcal{T}}\;
\mathcal{F}_{\text{FGW}}(\mathbf{C}, \mathbf{T})
\;+\; \gamma\, \mathrm{D}_{\mathrm{KL}}\!\big(\mathbf{T}\,^\top \mathbf{1}_K \,\big\|\, p_s\big),
\end{equation}
where $\mathcal{T} := \{\mathbf{T}\mid
\mathbf{T}^\top \mathbf{1}_{F}=\frac{1}{K}\mathbf{1}_K\}$ enforces balanced anchor marginals and relaxed frame marginals. 
The KL term in~\Cref{eq:fugw_prob}, controlled 
by $\gamma$, encourages the frame marginal $\mathbf{T}\mathbf{1}_K$ to follow the saliency distribution $p_s$. 
$\mathbf{C} = \{C^k, C^v, C^a\}$ represents the costs used in ASOT. Specifically, $C^v$ encodes an intra-segment temporal cost and $C^a$ captures an inter-segment semantic cost. 

{Second}, we incorporate $p_s$ as a 
bias term in the KOT cost matrix $C^k\in\mathbb{R}^{F\times K}$~\cite{KOT} to prioritize salient frames during the 
assignment process:
\begin{equation}\label{eq:cost_matrix}
C^k_{nj} := (1 - \frac{{x^{s}_n}^\top {a}_j}{\|{x}^s_n\|_2\|{a}_j\|_2}) - \mu p_{s_n},
\end{equation}
where $\mu$ controls the strength of the saliency bias, $x^{s}_n$ denotes spatial embeddings.

This dual use of saliency scores is realized through the optimized transport plan
$\mathbf{T}^\star \in \mathbb{R}^{F \times K}$, which allocates higher transport mass to important frames and suppresses background frames. This encourages the formation of segments centered on salient regions, resulting in semantically coherent clusters. Using $\mathbf{T}^\star$, each anchor $a_j$ is associated with the frames that receive the highest transport mass from that anchor. Adjacent frames assigned to the same anchor form a temporal segment $s_j = [f_j^s, f_j^e)$. However, balanced anchors may still produce segments containing non-salient frames. To exclude such uninformative segments, we assign each $s_j$ a score that captures both its semantic alignment and temporal stability:
\begin{equation}
\mathcal{S}_{\text{OT}}(s_j)
= \frac{1}{L_j} \sum_{i = f_j^s}^{f_j^e - 1} T^\star_{i,\,a_j}, 
\quad
\mathcal{S}_{\text{len}}(s_j)
= \log(1 + L_j),
\label{eq:scoring}
\end{equation}
where $L_j = f_j^e - f_j^s$ denotes the segment length.
The alignment score $\mathcal{S}_{\text{OT}}(s_j)$ measures the average 
transport mass assigned to anchor $a_j$, rewarding segments whose frames 
exhibit strong correspondence with their anchor. 
The length regularizer $\mathcal{S}_{\text{len}}(s_j)$ encourages 
segments with sufficient duration to maintain temporal consistency.
The final score is computed as
$\mathcal{S}(s_j) = \mathcal{S}_{\text{OT}}(s_j) \times \mathcal{S}_{\text{len}}(s_j)$,
which prioritizes segments that are both semantically aligned with their anchor 
and temporally stable.

We rank all segments by $\mathcal{S}(s_j)$ and keep the top-$k$ as retrieval units, resulting in $N_{\text{seg}}=k$ representative segments.
Each selected segment is represented by aggregating its frame embeddings, 
where we apply a saliency-weighted average to emphasize informative frames. 
This saliency-weighted pooling ensures that more salient frames have 
greater influence on retrieval. Using these $k$ representative features, 
we perform retrieval following the procedure described in \Cref{sec:preliminaries}
to obtain $R \in \mathbb{R}^{k\times D}$.

\paragraph{Saliency Prompt.}
\label{sec:Salip}
\textit{SaliP} converts frame-level saliency scores into prompts that guide the decoder toward salient moments. Saliency score $P_s$ is projected through a learnable mapping layer, implemented as a linear layer, to yield saliency prompts $S$. Let $X'$ denote the refined video frame embeddings, $R$ the retrieval embeddings, and $Y$ the transcribed text tokens. These components are concatenated into a unified sequence:
\begin{equation}
T_{in} = [X'; S; R; Y] \in \mathbb{R}^{(F + F + k + N_t) \times D}.
\end{equation}
Through \textit{SaliP}, saliency information is directly incorporated into decoding. This prompt explicitly guides temporal grounding, helping the decoder focus on salient frames during caption generation.

\subsection{Training and Inference}
\begin{table*}[t]
\centering
\resizebox{0.9\linewidth}{!}{
\begin{tabular}{l|c|cccc|cccc}
\toprule[2pt]
\multirow{2}{*}{\textbf{Method}} & \multirow{2}{*}{\textbf{PT}} & \multicolumn{4}{c|}{\textbf{YouCook2 (val)}} & \multicolumn{4}{c}{\textbf{ViTT (test)}} \\
\cmidrule(lr){3-6} \cmidrule(lr){7-10}
& & \textbf{CIDEr} & \textbf{METEOR} & \textbf{SODA\_c} & \textbf{BLEU\_4} & \textbf{CIDEr} & \textbf{METEOR} & \textbf{SODA\_c} & \textbf{BLEU\_4} \\
\midrule
PDVC~\pub{ICCV21} & \ding{55} & 29.69 & 5.56 & 4.92 & 1.40 & - & - & - & - \\
CM$^2$~\pub{CVPR24} & \ding{55} & 31.66 & 6.08 & 5.34 & 1.63 & - & - & - & - \\
E$^2$DVC~\pub{CVPR25} & \ding{55} & 34.26 & 6.11 & 5.39 & 1.68 & - & - & - & - \\
CACMI~\pub{AAAI26} & \ding{55} & 34.83 & 6.21 & 5.57 & 1.70 & - & - & - & - \\
\midrule
Streaming V2S~\pub{CVPR24} & \ding{51} & 32.90 & 7.10 & 6.00 & - & 25.20 & 5.80 & 10.00 & - \\
DIBS~\pub{CVPR24} & \ding{51} & 44.44 & 7.51 & 6.39 & - & - & - & - & - \\
Vid2Seq$^\dagger$~\pub{CVPR23} & \ding{51} & 66.29 & 12.41 & 9.87 & 5.64 & 48.84 & 9.51 & 14.99 & 0.71 \\
HiCM$^2$~\pub{AAAI25} & \ding{51} & {71.84} & {12.80} & \textbf{10.73} & 6.11 & {51.29} & {9.66} & {15.07} & {0.86} \\
Sali4Vid~\pub{EMNLP25} & \ding{51} & \underline{75.80} & \underline{13.54} & \underline{10.28} & \underline{6.35} & \underline{53.87} & \underline{10.05} & \underline{15.08} & \textbf{0.91} \\
\rowcolor{Grey} Ours & \ding{51} & \textbf{80.53} & \textbf{13.86} & \textbf{10.73} & \textbf{6.75} & \textbf{56.04} & \textbf{10.49} & \textbf{15.01} & \underline{0.89} \\
\bottomrule[2pt]
\end{tabular}}
\caption{Comparison with state-of-the-art methods on the YouCook2 validation set and the ViTT test set. PT indicates whether the model is pretrained. Bold and underlined values represent the best and second best results, respectively. ``--'' denotes unavailable results, and $\dagger$ marks results reproduced from official implementations. Our method achieves state-of-the-art performance across most evaluation metrics.}
\label{tab:main}
\end{table*}

\paragraph{Training.}
During training, the decoder takes $T_{\text{in}}$ as input, where 
the corresponding video frame features are the original features $X$. We use 
these raw frame-level features to enable precise alignment with text transcriptions. 
Following prior work~\cite{yang2023vid2seq, kim2025hicm2, jeon2025sali4vid}, the decoder is built upon the T5 architecture~\cite{raffel2020exploring}.
The T5 decoder is trained with a token-level Cross-Entropy (CE) loss between 
the predicted caption $z$ and ground truth captions $\hat{z}$. As described in ~\Cref{sec:Supervised Saliency Learning}, 
we also apply a saliency loss scaled by $\lambda$ to supervise the HD module. 
The final training objectives are defined as:
\begin{equation}
    \mathcal{L}_{\text{CE}} = \text{CE}(z, \hat{z}), \quad
    \mathcal{L}_{\text{total}} = \mathcal{L}_{\text{CE}} + \lambda \mathcal{L}_{\text{saliency}}.
\end{equation}

\paragraph{Inference.}
At inference, the decoder similarly takes $T_{\text{in}}$ as input, 
but the corresponding video frame features are the refined features $X'$ from 
~\Cref{sec:SWSA}. This design uses raw features at training time for fine-grained text alignment, 
while refined features at inference time provide richer temporal context, leading to more accurate event boundaries and coherent captions. Note that our method is formulated from the inference perspective, and the overall \textit{STaRC} pipeline follows this inference procedure.

\begin{table}[t]
    \centering
    \resizebox{\linewidth}{!}{ 
        \begin{tabular}{l|c|ccc|ccc}
        \toprule[2pt]
        \multirow{2}{*}{\textbf{Method}} & \multirow{2}{*}{\textbf{PT}} & \multicolumn{3}{c|}{\textbf{YouCook2 (val)}} & \multicolumn{3}{c}{\textbf{ViTT (test)}} \\
        \cmidrule(lr){3-5} \cmidrule(lr){6-8}
        & & \textbf{F1} & \textbf{Recall} & \textbf{Precision} & \textbf{F1} & \textbf{Recall} & \textbf{Precision} \\
        \midrule
        PDVC & \ding{55} & 26.81 & 22.89 & 32.37 & - & - & - \\
        CM$^2$ & \ding{55} & 28.43 & 24.76 & 33.38 & - & - & - \\
        E$^2$DVC~\pub{CVPR25} & \ding{55} & 28.87 & 25.01 & 34.13 & - & - & - \\
        CACMI~\pub{AAAI26} & \ding{55} & 29.34 & 25.54 & 34.63 & - & - & - \\
        \midrule
        Streaming V2S & \ding{51} & 24.10 & - & - & 35.40 & - & -  \\
        DIBS & \ding{51} & 31.43 & 26.24 & \textbf{39.81} & - & - & -  \\
        Vid2Seq$^\dagger$ & \ding{51} & 31.08 & 30.38 & 31.81 & \underline{46.21} & \textbf{45.89} & 46.53 \\
        HiCM$^2$ & \ding{51} & 32.51 & \textbf{32.51} & 32.51 & 45.98 & \underline{45.00} & {47.00} \\
        Sali4Vid & \ding{51} & \underline{33.61} & {31.11} & {36.57} & \textbf{46.58} & 44.31 & \underline{49.10} \\
        \rowcolor{Grey}
        \textbf{Ours} & \ding{51} & \textbf{34.34} & \underline{31.53} & \underline{37.70} & {44.34} & 40.02 & \textbf{49.72} \\
        \bottomrule[2pt]
    \end{tabular}}
    \caption{
    Localization results on the YouCook2 validation set and the ViTT test set. Bold and underlined values indicate the best and second-best performances, respectively. ``--'' denotes unavailable results.}
    \label{tab:localization}
\end{table}

\section{Experiments}
\label{sec:experiments}
\subsection{Experimental Setup} 
\paragraph{Datasets and Evaluation Metrics.}
We conduct experiments on two Dense Video Captioning (DVC) benchmarks, {YouCook2}~\cite{youcook2} and {ViTT}~\cite{vitt}. 
The YouCook2 dataset contains 2,000 untrimmed cooking videos, each averaging 320 seconds in length and annotated with 7.7 temporally localized sentences. The ViTT dataset consists of 8,000 untrimmed instructional videos, averaging 250 seconds per video, and annotated with 7.1 temporally localized short tags. We evaluate our model on two sub-tasks within DVC, focusing on caption generation and event localization. Following the official evaluation protocol~\cite{wang2020dense}, we measure caption quality using CIDEr~\cite{vedantam2015cider}, METEOR~\cite{banerjee2005meteor}, and BLEU\_4~\cite{papineni2002bleu}, which compare generated captions with ground truth across IoU thresholds of 0.3, 0.5, 0.7, and 0.9. To further assess storytelling ability, we employ the SODA\_c metric~\cite{fujita2020soda}.
For event localization, we compute the average precision (AP), average recall (AR), and F1 score, averaged across the same IoU thresholds.
\vspace{-5pt}

\paragraph{Implementation Details.}
Following prior studies~\cite{kim2025hicm2, jeon2025sali4vid}, our framework is built upon Vid2Seq~\cite{yang2023vid2seq}, pre-trained on 1.8M video–text pairs.
Video frames are uniformly sampled at 1 FPS and truncated or zero-padded to a fixed sequence length of $T=100$. We adopt the two-stage training protocol from Sali4Vid~\cite{jeon2025sali4vid}.
First, the model is pre-trained following the original Vid2Seq configuration with a learning rate of 3e-4.
Then, we fine-tune the model for 10 additional epochs using our proposed method.
During fine-tuning, the learning rate starts at 1e-5, is linearly warmed up over the first 10\% of training steps, and is subsequently decayed to zero with a cosine schedule. All experiments are conducted on a single NVIDIA A6000 GPU with a batch size of 4. For the saliency-aware loss, we set $\lambda = 6.0$ for YouCook2 and $\lambda = 2.0$ for ViTT.
The \textit{Sliding Window Self-Attention (SWSA)} module operates with window sizes of 8, 32, and 64.
In the \textit{Saliency-Guided Segmentation and Retrieval (SGSR)} stage, we use 8 anchors to define 8 candidate segments and select the top-5 segments with the highest scores.
The hyperparameters $\mu$ and $\rho$ are set to 0.1 and 0.3, respectively.
Other segmentation components follow the default configuration of ASOT~\cite{xu2024temporally}.
Ablation studies and additional analyses of the segmentation mechanism are provided in the Appendix.





    


\begin{table}[t]
\centering
\resizebox{0.90\linewidth}{!}{
\begin{tabular}{l|cccc}
\toprule[2pt]

    Method  & SGSR \ref{sec:SGSR}&SaliP \ref{sec:Salip}& CIDEr & METEOR \\
    \midrule
    \textit{Baseline} &&& 66.29 & 12.41 \\
    \midrule

    \textit{(1)} &\checkmark&& 76.94 & 13.60 \\
    
    \textit{(2)} &&\checkmark&  {78.74} & {13.75} \\


    \midrule
    \textit{STaRC} &\checkmark&\checkmark&  \textbf{80.53} & \textbf{13.86} \\
    \bottomrule[2pt]
  \end{tabular}}
  \caption{Ablation study on the key components of \textit{STaRC}.
We evaluate the contributions of our main proposed modules, \textit{SGSR}, and \textit{SaliP} on YouCook2 validation set. The baseline corresponds to the performance of Vid2Seq.}
  \label{tab:module_ablation}
\end{table}

\begin{table}[t]
\centering
\resizebox{0.99\linewidth}{!}{
\begin{tabular}{l|cccc}
\toprule[2pt]
\multirow{2}{*}{\textbf{\shortstack{Window sizes}}} & \multicolumn{4}{c}{\textbf{YouCook2 (val)}} \\
\cmidrule(lr){2-5} 
&\textbf{CIDEr} & \textbf{METEOR} & \textbf{SODA\_c} & \textbf{BLEU\_4} \\
\midrule
w/o \textit{SWSA} ~\ref{sec:SWSA}   &75.82	&13.23	&10.33	&6.49\\
\midrule
6, 36, 66   &78.46	&13.66	&10.57	&6.66\\
\rowcolor{Grey}8, 32, 64 &\textbf{80.53}&\textbf{13.86}&\underline{10.73}&\underline{6.75}\\

10, 40, 70  &70.52	&13.26	&9.72	&5.92\\
\midrule
8   &78.03	&13.69	&10.62	&6.65\\
8, 32   &\underline{79.35}	&\underline{13.80}	&\textbf{10.79}	&6.73\\
8, 16, 32, 64   &{79.06}	&{13.75}	&{10.62}	&\textbf{6.78}\\
\bottomrule[2pt]
\end{tabular}}
\caption{Ablation study on window sizes and the effect of removing \textit{SWSA} module on the YouCook2 validation set.}
\label{tab:stride_ablation}
\end{table}
\subsection{Results and Analysis}
\label{sec:experiment}
\paragraph{Comparison with State-of-the-Art.}
\Cref{tab:main} and \Cref{tab:localization} compare \textit{Saliency Training for Retrieval and Captioning (STaRC)} with recent DVC methods. \textit{STaRC} attains a CIDEr score of \textbf{80.53} on YouCook2 and \textbf{56.04} on ViTT, outperforming prior methods such as Sali4Vid~\cite{jeon2025sali4vid} and HiCM$^2$~\cite{kim2025hicm2}.
For localization, \textit{STaRC} achieves the best F1 score of 34.34 on YouCook2 and the highest precision of 49.72 on ViTT.
Unlike previous retrieval-augmented models that rely on heuristic segmentation or timestamp-derived saliency, \textit{STaRC} benefits from supervised saliency signals that produce event-aligned segments.
These segments also supply the decoder with more reliable temporal cues during generation, leading to more faithful event descriptions.
\vspace{-5pt}
\paragraph{Component Ablation Study.}

We validate the contribution of each proposed key module \textit{SGSR} and \textit{SaliP}. As shown in~\Cref{tab:module_ablation}, combining both components yields the best performance, achieving a +14.24 CIDEr gain over the baseline Vid2Seq model. When applied individually, \textit{SaliP} alone brings a +12.45 CIDEr improvement, while \textit{SGSR} alone yields +10.65. Notably, both modules provide meaningful gains on their own, and their combination further boosts performance, showing that they are complementary. This confirms the effectiveness of our \textit{Unified Saliency-Based Design}, where segmentation and caption generation both benefit from the same learned saliency signal.
\begin{figure*}[t]
\begin{center}
\includegraphics[width=1\textwidth]{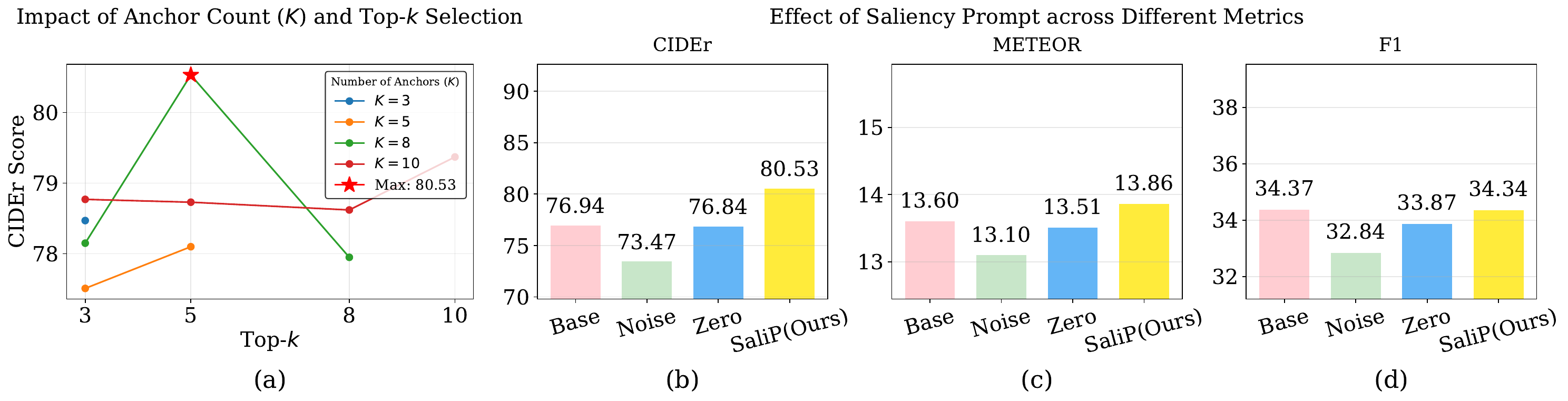}
\end{center}
\vspace{-12pt}
\caption{
\textbf{(a) Effect of anchor count and top-$k$ segment selection.}
CIDEr scores across different numbers of anchors ($K$) and top-$k$ segment selections, where only the $k$ highest-scoring segments are used for retrieval. Each line represents a different anchor configuration, and $K$ corresponds to the number of segments produced by \textit{SGSR}.  
\textbf{(b-d) Impact of saliency-prompt corruption.}
These plots show the effect of replacing \textit{SaliP} with Zero and Gaussian noise prompts. Both corrupted prompts still outperform the baseline (\textit{STaRC} without \textit{SaliP}), but Gaussian noise causes a larger drop, emphasizing the importance of accurate saliency cues.
}
\label{fig:combined_results}
\end{figure*}
\vspace{-5pt}
\paragraph{Feature Refinement Ablation Study.}

We analyze the effect of our feature refinement module, Sliding-Window Self-Attention (\textit{SWSA}). As presented in~\Cref{tab:stride_ablation}, removing it from the full model results in a slight drop in CIDEr, indicating that this module provides a complementary gain when combined with the two key modules. We also explore different window size configurations. To ensure consistent temporal coverage, we employ three windows, each defined as a multiple of the smallest one. Among the tested configurations, the combination of window sizes 8, 32, and 64 with three windows achieves the best performance.

\begin{table}[t]
\centering
\resizebox{0.99\linewidth}{!}{
\begin{tabular}{l|cccc}
\toprule[2pt]
\multirow{2}{*}{\textbf{Method / Variant}} & \multicolumn{4}{c}{\textbf{YouCook2 (val)}} \\
\cmidrule(lr){2-5}
& \textbf{CIDEr} & \textbf{METEOR} & \textbf{SODA\_c} & \textbf{BLEU\_4} \\
\midrule
\multicolumn{5}{l}{\textit{\textbf{Clustering Type}}} \\
\midrule
k-means & 75.63 & 13.34 & 10.46 & 6.29 \\
Adaptive clustering~\cite{jeon2025sali4vid} & \underline{78.19} & \underline{13.69} & \underline{10.55} & \underline{6.57} \\
 \rowcolor{Grey}SGSR (Ours) &\textbf{80.53}& \textbf{13.86}& \textbf{10.73}& \textbf{6.75}\\
\midrule
\multicolumn{5}{l}{\textit{\textbf{Pooling Strategy (within SGSR)}}} \\
\midrule
Uniform Avg & 78.95 & 13.65 & 10.52 & 6.68 \\
 \rowcolor{Grey}Saliency-weighted Avg (Ours)&\textbf{80.53}& \textbf{13.86}& \textbf{10.73}&\textbf{6.75}\\
\bottomrule[2pt]
\end{tabular}}
\caption{Ablation on segmentation and pooling strategies. The top part compares clustering methods, where our \textit{SGSR} outperforms k-means and adaptive clustering~\cite{jeon2025sali4vid}.
The bottom part compares pooling strategies within \textit{SGSR}, showing the benefit of saliency-weighted averaging over uniform averaging.}

\label{tab:cluster_ablation}
\end{table}

\begin{table}[t]
\centering
\resizebox{0.95\linewidth}{!}{
\begin{tabular}{l|cccc}
\toprule[2pt]
\multirow{2}{*}{\textbf{\shortstack{Num of\\Retrieval ($p$)}}} & \multicolumn{4}{c}{\textbf{YouCook2 (val)}} \\
\cmidrule(lr){2-5} 
&\textbf{CIDEr} & \textbf{METEOR} & \textbf{SODA\_c} & \textbf{BLEU\_4} \\
\midrule
3   &\underline{79.65}&\underline{13.79}&10.61&\textbf{6.79}\\
5   &79.12&13.73&\underline{10.70}&6.67\\
\rowcolor{Grey}10  &\textbf{80.53}&\textbf{13.86}&\textbf{10.73}&\underline{6.75}\\
15  &78.00&13.81&10.56&6.57\\
\bottomrule[2pt]
\end{tabular}}
\caption{Ablation study on the number of top-ranked (top-$p$) captions retrieved based on similarity.}
\label{tab:retrievalk_ablation}
\end{table}

\vspace{-5pt}
\paragraph{Saliency-Guided Segmentation and Retrieval Ablation Study.}
To verify the contribution of our \textit{SGSR}, we compare it with alternative segmentation and pooling strategies in~\Cref{tab:cluster_ablation}.
Our method achieves +4.9 and +2.34 CIDEr gains over standard k-means and the adaptive clustering scheme from Sali4Vid~\cite{jeon2025sali4vid}, respectively. We also find that constructing segment representations with saliency-weighted averaging, instead of uniform averaging, further improves performance. The learned saliency scores emphasize informative frames when forming segments, resulting in more relevant retrieval. We further analyze how varying the number of anchors ($K$) and retrieved segments ($k$) affects overall performance, as shown in~\Cref{fig:combined_results} (a).
Additional analysis and hyperparameter details are provided in the Appendix. In~\Cref{tab:retrievalk_ablation}, we compare different numbers of retrieved captions per segment ($p \in \{3, 5, 10, 15\}$).
Retrieving 3 or 5 captions results in noticeably lower performance across all metrics. In contrast, retrieving 15 captions does not lead to further improvement and instead causes a decline in performance. We observe that retrieving 10 captions provides the highest scores across most metrics, consistent with prior work~\cite{jeon2025sali4vid, kim2025hicm2}.

\vspace{-5pt}
\paragraph{Saliency Prompt Ablation Study.}
To investigate the contribution of \textit{SaliP}, we conduct an ablation study in~\Cref{fig:combined_results} (b-d) by replacing the saliency prompt with two variants: a prompt with added Gaussian noise and a Zero prompt of the same shape. The baseline corresponds to \textit{STaRC} without the \textit{SaliP} module. Both variants outperform the baseline, indicating that additional tokens help the decoder capture temporal and contextual information. However, prompt quality is crucial. The Gaussian noise prompt causes the largest performance drop. It corrupts 
saliency scores with random noise, actively misleading the decoder to focus on incorrect frames. The Zero prompt performs better by assigning uniform importance to all frames. This avoids misleading cues but also removes temporal guidance. Without saliency information, the decoder cannot distinguish important moments from background, reducing caption quality. This sensitivity to prompt quality demonstrates that the model relies on accurate saliency cues from \textit{SaliP} for effective DVC.

\subsection{Qualitative Results}
\begin{figure}[t]
    \centering
    \includegraphics[width=0.99\linewidth]{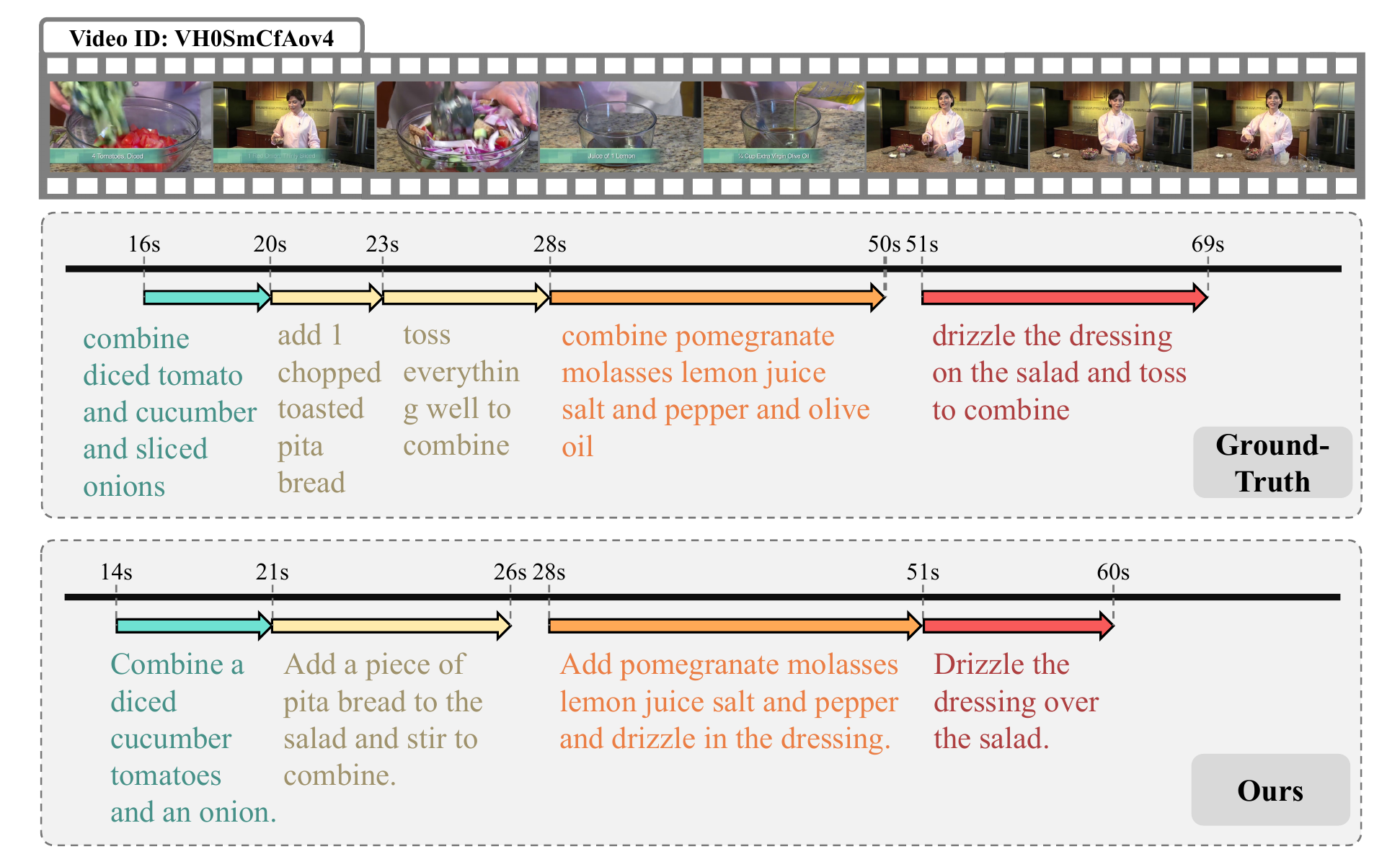}
    \caption{
    \textbf{A qualitative result from YouCook2 validation set.} The arrows indicate the duration of each localized event, and the text below each arrow represents its corresponding caption.}
\label{fig:DVC_qual}
\vspace{-5pt}
\end{figure}

As shown in \Cref{fig:DVC_qual}, which illustrates an example from the YouCook2 validation set, our proposed \textit{STaRC} accurately predicts event durations and generates captions that align well with the ground truth descriptions. The predicted segments also exhibit clear temporal boundaries with coherent correspondence between visual content and textual semantics. Notably, \textit{STaRC} is able to capture subtle transitions between consecutive events, producing more precise segment splits.

\section{Conclusion}

We propose \textit{STaRC}, a unified saliency-guided framework for retrieval-augmented 
Dense Video Captioning. By learning frame-level saliency directly from event 
annotations, \textit{STaRC} provides a temporal signal that guides both retrieval 
and captioning. \textit{SGSR} produces event-aligned segments via Optimal Transport, 
and \textit{SaliP} injects saliency into the decoder as temporal prompts. Experiments 
on YouCook2 and ViTT demonstrate that \textit{STaRC} achieves strong performance, 
validating supervised saliency as an effective signal.

\paragraph{Acknowledgement}
This was partly supported by the Institute of Information \& Communications Technology Planning \&
Evaluation (IITP) grant funded by the Korean government (MSIT) (No.RS-2020-II201373, Artificial
Intelligence Graduate School Program(Hanyang
University)) and the Institute of Information \&
Communications Technology Planning \& Evaluation (IITP) grant funded by the Korean government (MSIT) RS-2025-25422680, Metacognitive
AGI Framework and its Applications).


{
    \small
    \bibliographystyle{ieeenat_fullname}
    \bibliography{main}
}

\appendix
\clearpage
\setcounter{page}{1}
\setcounter{table}{0}
\setcounter{figure}{0}
\maketitlesupplementary
\renewcommand{\thetable}{A.\arabic{table}}
\renewcommand{\thefigure}{A.\arabic{figure}}

In this Appendix, we present additional experiments, detailed formulations, and qualitative results to further support our findings on the proposed \textit{STaRC}. Specifically, \S\ref{supp:sup_training} provides additional analyses and experiments on \textit{Supervised Saliency Training}, \S\ref{supp:retrieval} presents extended formulations and studies for \textit{Saliency-Guided Segmentation and Retrieval (SGSR)}, and \S\ref{supp:add_exp} additional qualitative results and visualizations, particularly focusing on experiments regarding ~\textit{Sliding Window Self-Attention (SWSA)} module.

\section{Supervised Saliency Training}
\label{supp:sup_training}
\S\ref{supp:loss_hyp} analysis the impact of hyperparameters ($\lambda$, $\tau$) in the saliency loss and \S\ref{supp:qualitative} offers visualizations of predicted saliency scores.

\subsection{Saliency Loss Hyperparameters}
\label{supp:loss_hyp}
\begin{figure}[t]
    \centering
    \includegraphics[width=0.9\linewidth]{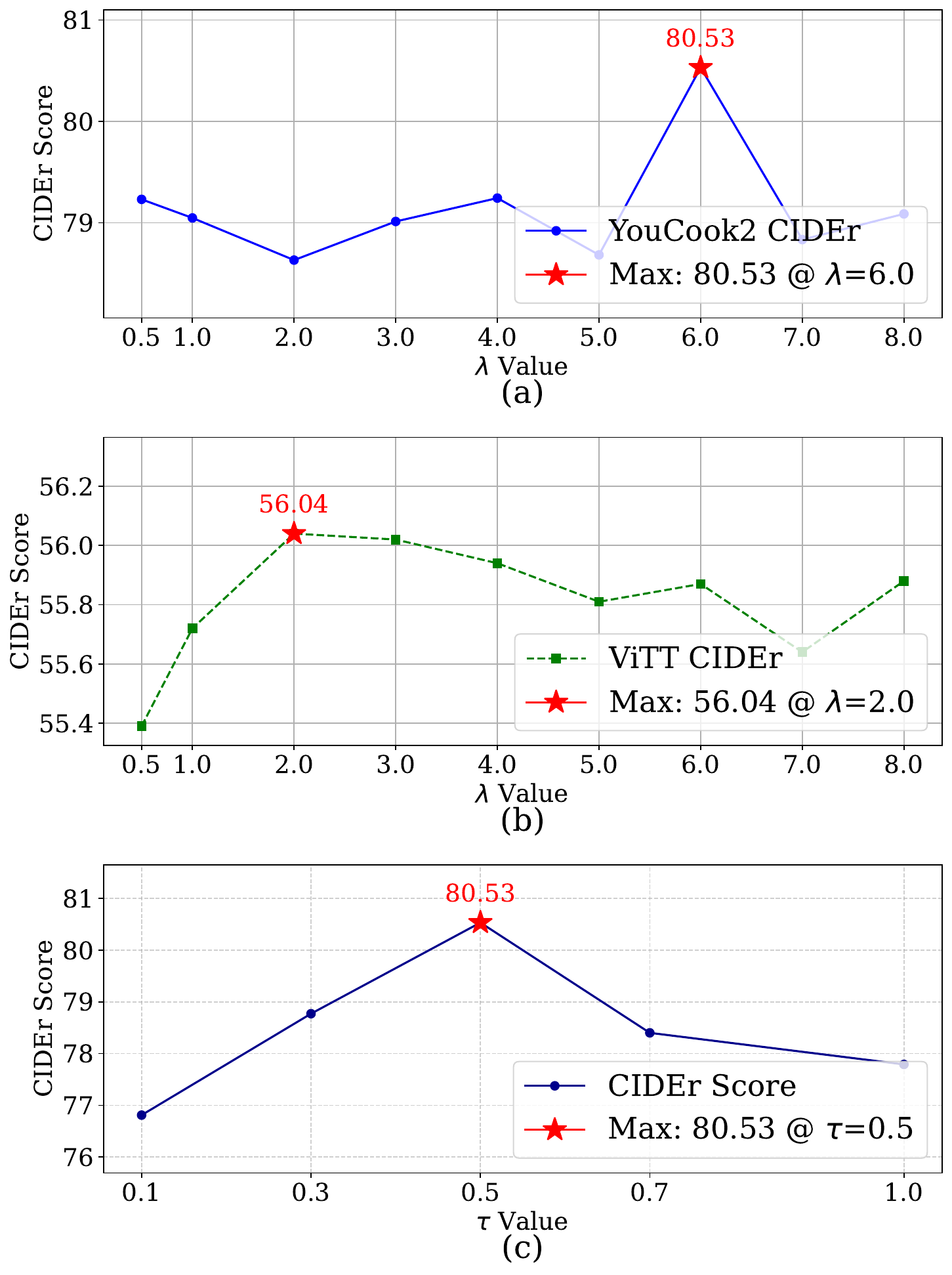}
    \caption{
\textbf{(a-b) CIDEr performance for different saliency-loss weights $\lambda$.}
Results on YouCook2 and ViTT showing how different values of $\lambda$ influence the strength of supervised saliency learning and overall caption quality. 
\textbf{(c) CIDEr performance in YouCook2 validation set for different temperature values $\tau$.}
This figure visualizes model sensitivity to different $\tau$ values used in the loss formulation.
}
\label{supp:lambda_tau}
\end{figure}

The hyperparameters $\lambda$ and $\tau$ significantly impact saliency learning performance, as shown in~\Cref{supp:lambda_tau}. 
These hyperparameters appear in our saliency loss formulation (\Cref{eq:softmax_prob_single} and~\Cref{eq:video_loss_single}), 
where $\tau$ controls the softmax temperature and $\lambda$ weights the saliency loss. YouCook2~\cite{youcook2} achieves best performance at $\lambda = 6.0$, while ViTT~\cite{vitt} performs 
best at $\lambda = 2.0$. For the temperature parameter $\tau$, $0.5$ provides optimal performance.

\subsection{Qualitative Analysis of Saliency Scores}
\label{supp:qualitative}
\begin{figure*}[t]
\begin{center}
\includegraphics[width=1\textwidth]{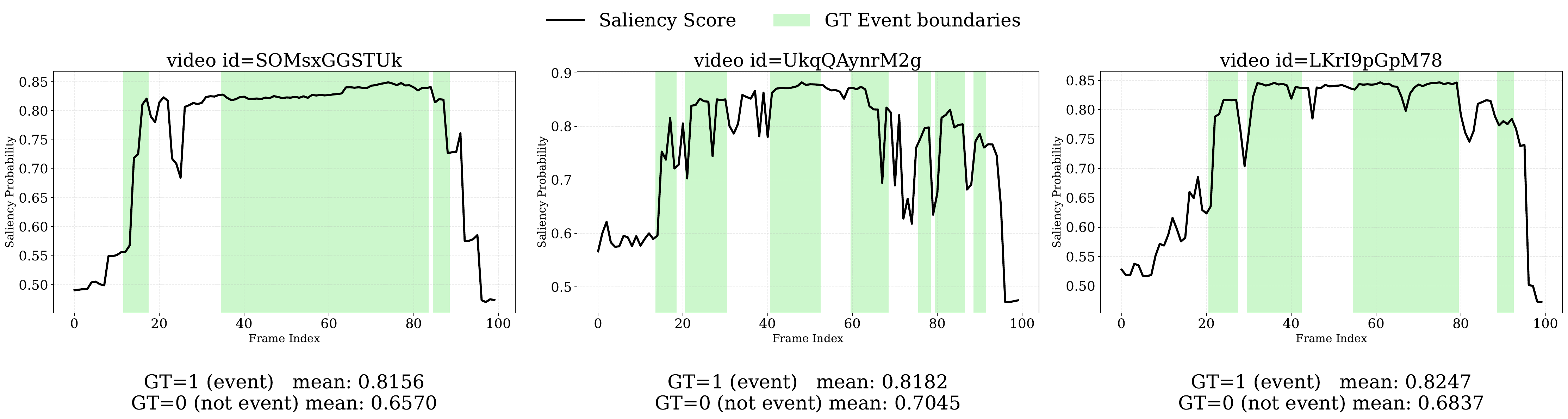}
\end{center}
\caption{\textbf{Visualization of predicted frame-level saliency scores.} Black lines show predicted saliency scores overlaid with ground truth event boundaries (green regions). Saliency scores are higher within event regions than in non-event regions. This separation confirms that the model successfully learns to align saliency prediction with true event boundaries.
}
\label{supp:saleincy_vis}
\end{figure*}

We visualize the predicted saliency scores from the trained highlight detection module in~\Cref{supp:saleincy_vis}. Saliency scores are high within ground truth event boundaries and low in non-event regions. Quantitative analysis confirms that event regions show significantly higher mean scores than non-event regions, demonstrating effective separation of important frames from background.

\section{Saliency-Guided Segmentation and Retrieval}
\label{supp:retrieval}
\S\ref{supp:OT_supp} provides detailed equations for the saliency-aware OT-based segmentation method in \textit{SGSR}, including Kantorovich OT matching, Gromov-Wasserstein OT, fused OT objective, and the balanced-unbalanced constraint that uniquely differentiates \textit{SGSR} from prior OT-based formulations such as ASOT~\cite{xu2024temporally}, along with hyperparameter experiments.
\S\ref{supp:ret_supp} presents the saliency weighted pooling equation and qualitative segmentation results with retrieved captions.

\paragraph{Preliminary.}
Given frame spatial features $X^s \in \mathbb{R}^{F \times D}$ and learnable anchors
$A = \{a_j\}_{j=1}^K \in \mathbb{R}^{K \times D}$, we solve for a soft assignment matrix
$T \in \mathbb{F}^{N \times K}$, where $T_{ij}$ denotes the probability
that frame $i$ belongs to prototype $j$.

\subsection{OT-based segmentation method}
\label{supp:OT_supp}
\paragraph{Saliency-aware Kantorovich OT matching.}

We formulate the OT problem using the Kantorovich objective~\cite{thorpeot}. Let the anchor and frame marginals be defined as $\mathbf{p} = \frac{1}{F}1_F$ and $\mathbf{q} = \frac{1}{K}1_K$:
\begin{equation}\label{eq:ot_prob}
    \begin{array}{cc}
       \underset{\mathbf{T}\in \mathcal{T}_{\mathbf{p}, \mathbf{q}}}{\text{minimize}} & \mathcal{F}_{\text{KOT}}(\mathbf{C}^k, \mathbf{T}) := \langle \mathbf{C}^k, \mathbf{T}\rangle.
    \end{array}
\end{equation}

The visual affinity between frames and anchors is measured using cosine similarity. To incorporate saliency information, we introduce a bias term weighted by $\mu$ using the saliency prior $p_s$.
This yields the following cost matrix:
\begin{equation}\label{eq:cost_matrix_supple}
C^k_{nj} := (1 - \frac{{x^{s}_n}^\top {a}_j}{\|{x}^s_n\|_2\|{a}_j\|_2}) - \mu p_{s_n}.
\end{equation}

\paragraph{Temporal consistency via Gromov-Wasserstein OT.}
\label{supp:temp_gw_ot}
To enforce smooth temporal structure, we adopt the Gromov-Wasserstein (GW) OT~\cite{icml_16_peyre_gw}.
Let $C^{v}$ capture temporal proximity between frames and $C^{a}$ captures the structural relationship between prototypes. We define $[F]:=\{1, ..., F\}$. The GW objective is then formulated as:
\begin{equation}\label{eq:gwot_obj}
    \mathcal{F}_{\text{GW}}(\mathbf{C}^v, \mathbf{C}^a, \mathbf{T}) := \sum_{\substack{n,F \in [F]\\j,K \in A}} L(C_{nF}^v, C_{jK}^a) T_{nj}T_{FK}.
\end{equation}

\paragraph{Fused GW Optimal Transport Objective.}
\label{supp:fused_gw_ot}
The final \textit{SGSR} objective combines visual matching and structural consistency as Fused GW OT objective:
\begin{equation}\label{eq:fgwot_obj}
    \mathcal{F}_{\text{FGW}}(\mathbf{C}, \mathbf{T}) := \alpha\mathcal{F}_{\text{GW}}(\mathbf{C}^v, \mathbf{C}^a, \mathbf{T}) + (1-\alpha) \mathcal{F}_{\text{KOT}}(\mathbf{C}^k, \mathbf{T}).
\end{equation}

\paragraph{Unbalanced OT.}
\label{supp:unbal_ot}
Unlike ASOT, we reverse the balanced/unbalanced configuration to better suit the DVC task. Specifically, we apply unbalanced OT for frame-level matching using saliency scores as the marginal constraint $\mathbf{p}$, while the anchor marginal remains balanced, as formulated in~\Cref{eq:fugw_prob_supple}:
\begin{equation}\label{eq:fugw_prob_supple}
\min_{\mathbf{T}\in\mathcal{T}_q}\;
\mathcal{F}_{\text{FGW}}(\mathbf{C}, \mathbf{T})
\;+\; \gamma\, \mathrm{D}_{\mathrm{KL}}\!\big(\mathbf{T}\,^\top \mathbf{1}_K \,\big\|\, p_s\big).
\end{equation}

\begin{figure}[t]
    \centering
\includegraphics[width=0.98\linewidth]{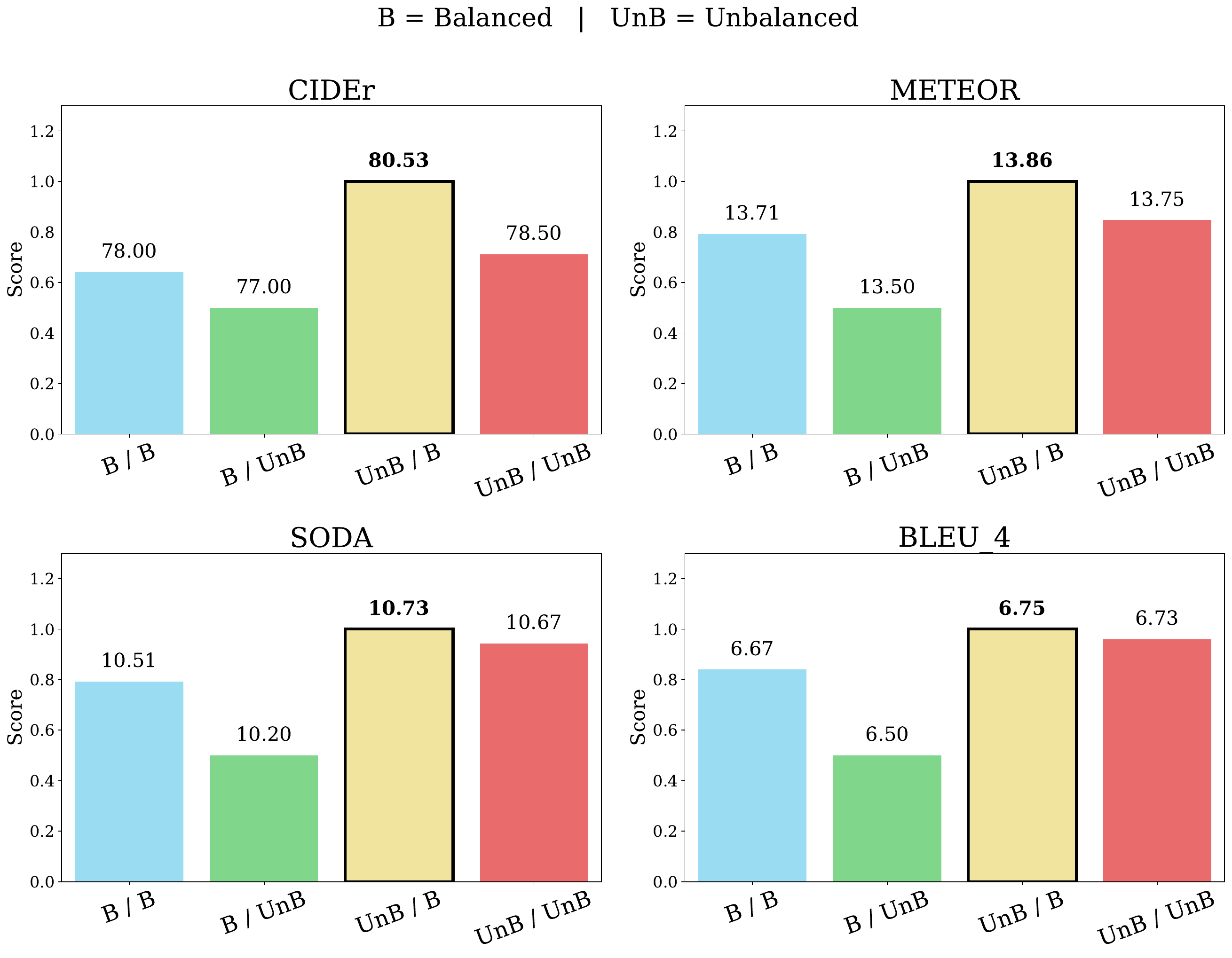}
\caption{\textbf{Performance comparison of \textit{SGSR} under different OT configurations.} We evaluate Balanced (B) and Unbalanced (UnB) OT for both frame-side and anchor-side matching. The configuration with UnB for frames and B for anchors achieves the best performance across all metrics.}
\label{supp:bal_unbal}
\end{figure}

This configuration enables the transport mass to follow the saliency distribution, allowing the model to adaptively allocate attention to salient regions. As shown in~\Cref{supp:bal_unbal}, this reversed configuration achieves the best performance for our task. Therefore, our ~\textit{SGSR} setting of applying unbalanced OT with saliency score constraints for frames and balanced OT for anchors proves to be optimal for DVC.

\begin{figure}[t]
    \centering
\includegraphics[width=0.98\linewidth]{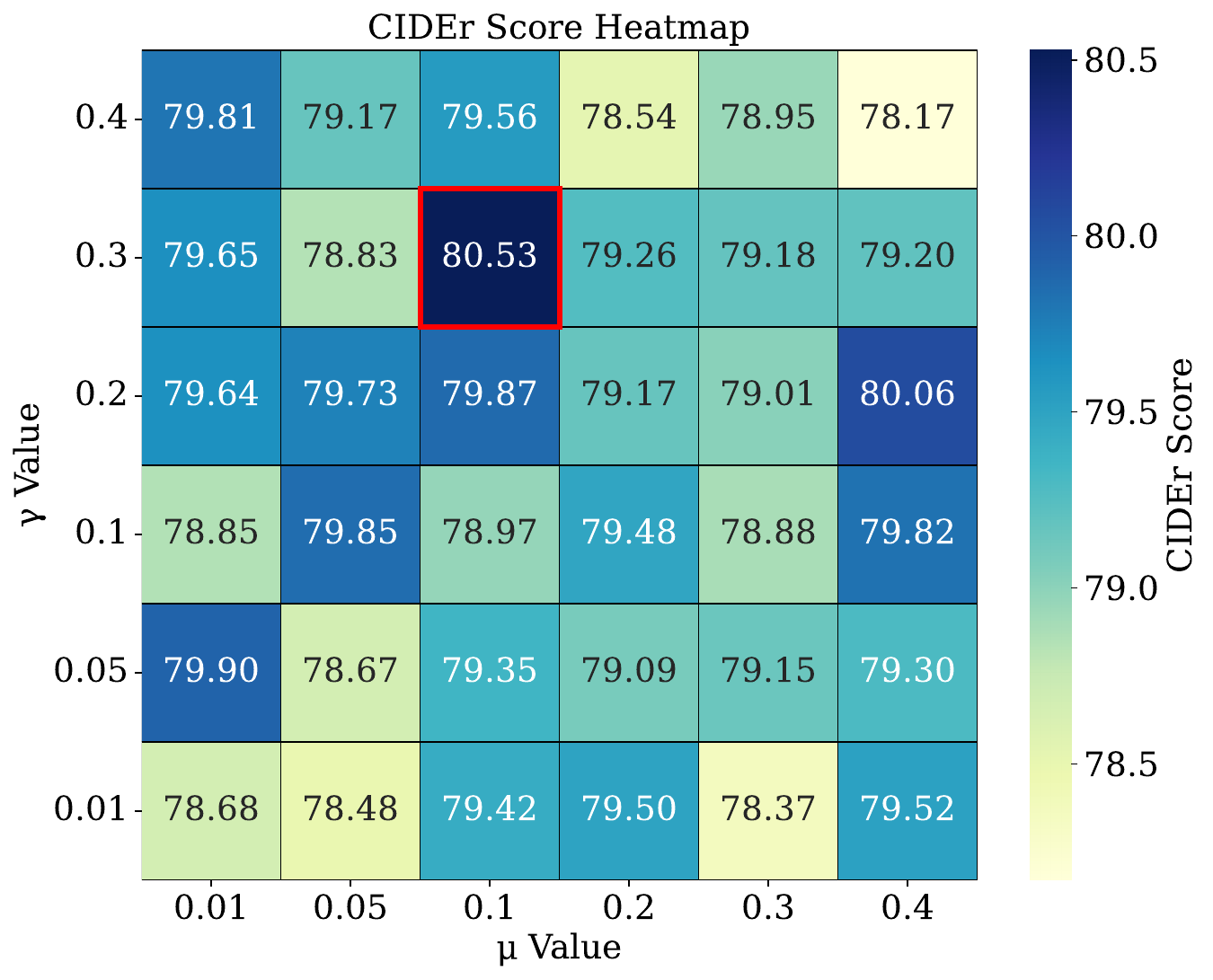}
\caption{
    \textbf{CIDEr performance heatmap across different combinations of saliency weight $\mu$ and temporal-consistency weight $\gamma$.} Each cell shows the CIDEr score obtained under a specific $\mu$ and $\gamma$ configuration in the \textit{SGSR} objective.
}
\label{supp:mu_gamma}
\end{figure}

\paragraph{Hyperparameter Analysis.}
\label{supp:ot_hyp}
Except for the saliency-related terms $\mu$ and $\gamma$, we keep all other OT parameters identical to ASOT. We set the unbalanced regularization coefficient $\rho$ to $0$ during both training and inference. This improves CIDEr from $79.61$ to $80.53$. We analyze different $\mu$ and $\gamma$ combinations in ~\Cref{supp:mu_gamma}. The best performance is achieved at $\mu = 0.1$ and $\gamma = 0.3$. This shows that moderate saliency weighting and temporal consistency work best for our task.
\begin{table}[t]
\centering
\resizebox{0.99\linewidth}{!}{
\begin{tabular}{l|cccc}
\toprule[2pt]
\multirow{2}{*}{\textbf{\shortstack{Datstore Type}}} & \multicolumn{4}{c}{\textbf{YouCook2 (val)}} \\
\cmidrule(lr){2-5} 
&\textbf{CIDEr} & \textbf{METEOR} & \textbf{SODA\_c} & \textbf{BLEU\_4} \\
\midrule
CC3M        &76.65&\underline{13.63}&10.53&6.63\\
COCO        &\underline{77.31}&13.57&\underline{10.66}&\underline{6.67}\\
Hierarchical~\cite{kim2025hicm2}&77.28&13.59&10.42&6.59\\
\rowcolor{Grey}In-domain&\textbf{80.53}&\textbf{13.86}&\textbf{10.73}&\textbf{6.75}\\
\bottomrule[2pt]
\end{tabular}}
\caption{Ablation study on different datstores used for retrieval.}
\label{tab:bank_ablation}
\end{table}

\subsection{Retrieval process}
\label{supp:ret_supp}
\paragraph{Representative Segment Features for Retrieval.}
After obtaining the optimized transport plan $T^*$, we generate segments. We rank all segments using the scoring and select the top-$k$ segments for retrieval. For each selected segment, we construct a representative feature by aggregating its frame embeddings using a saliency-weighted average to emphasize informative frames.

Let $s_j = [f_j^s, f_j^e)$ denote the frame indices assigned to segment $j$. The representative feature is computed as:
\begin{equation}\label{eq:sal_weight}
\bar{x}^s_j = \frac{\sum_{n \in s_j} p_{s,n} \, x^s_n}{\sum_{n \in s_j} p_{s,n}},
\end{equation}

These $k$ representative features are then used for retrieval following the mechanism described in~\Cref{sec:preliminaries}.
As shown in~\Cref{tab:cluster_ablation}, the saliency-weighted pooling leads to more accurate segment representations for retrieval. This results in consistent improvements in overall DVC performance compared to the uniform averaging strategy used in previous work~\cite{cm2,kim2025hicm2,jeon2025sali4vid}.

In addition, we evaluate retrieval performance under different datastores to assess the robustness of our method.
As reported in~\Cref{tab:bank_ablation}, \textit{SGSR} consistently maintains state-of-the-art performance.

\begin{figure}[t]
    \centering
    \includegraphics[width=1.0\linewidth]{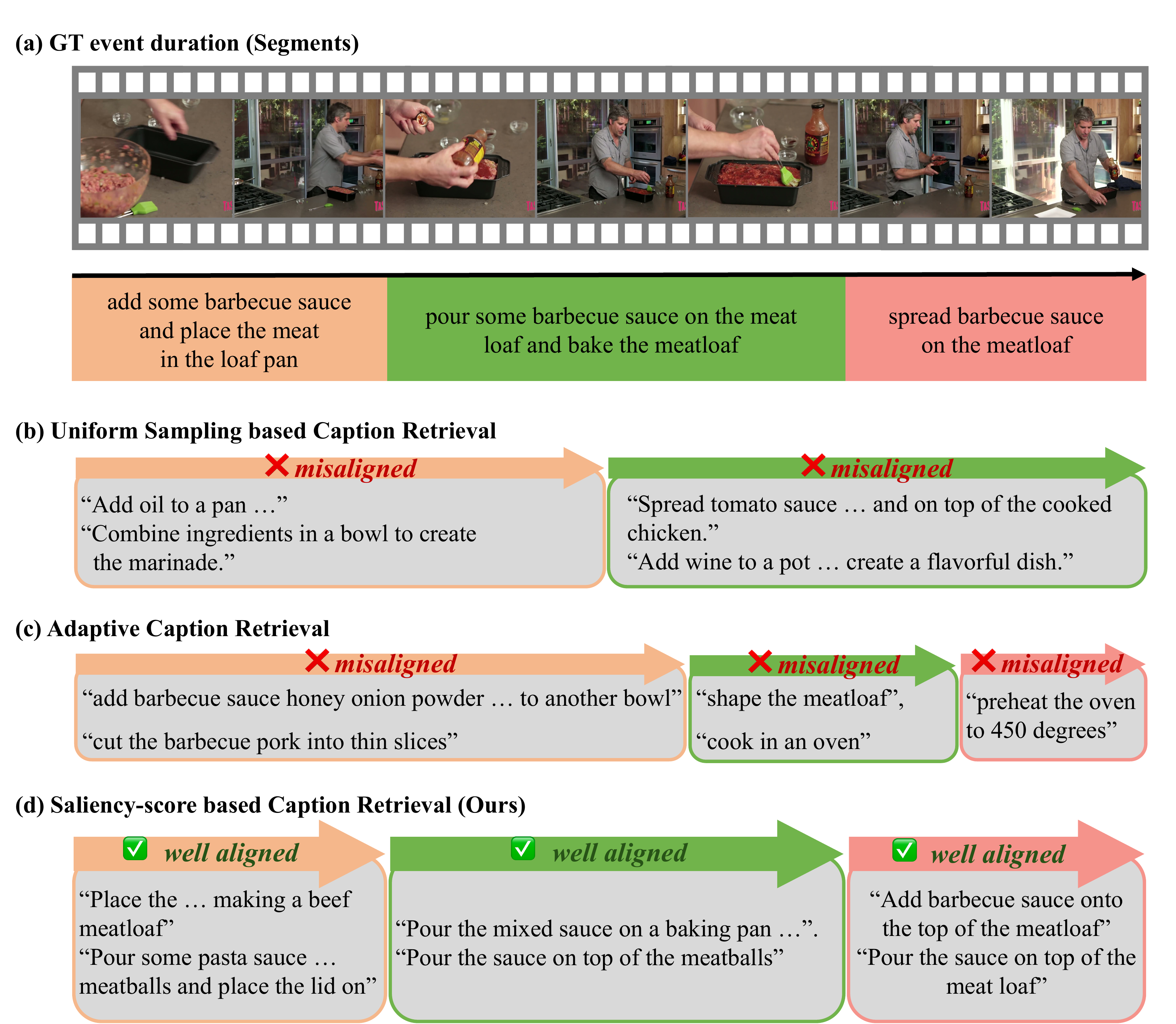}
\caption{
\textbf{Qualitative comparison of retrieved captions under different segmentation strategies.}
(a) Ground truth segments and captions.
(b) Segments from uniform sampling (HiCM2~\cite{kim2025hicm2}) and their retrieved captions.
(c) Segments from adaptive sampling (Sali4Vid~\cite{jeon2025sali4vid}) and their retrieved captions.
(d) Our \textit{SGSR} produces segments that align more closely with true event boundaries and retrieves more relevant captions.}
\label{supp:seg_qual}
\end{figure}

\begin{figure}[t]
    \centering
\includegraphics[width=0.98\linewidth]{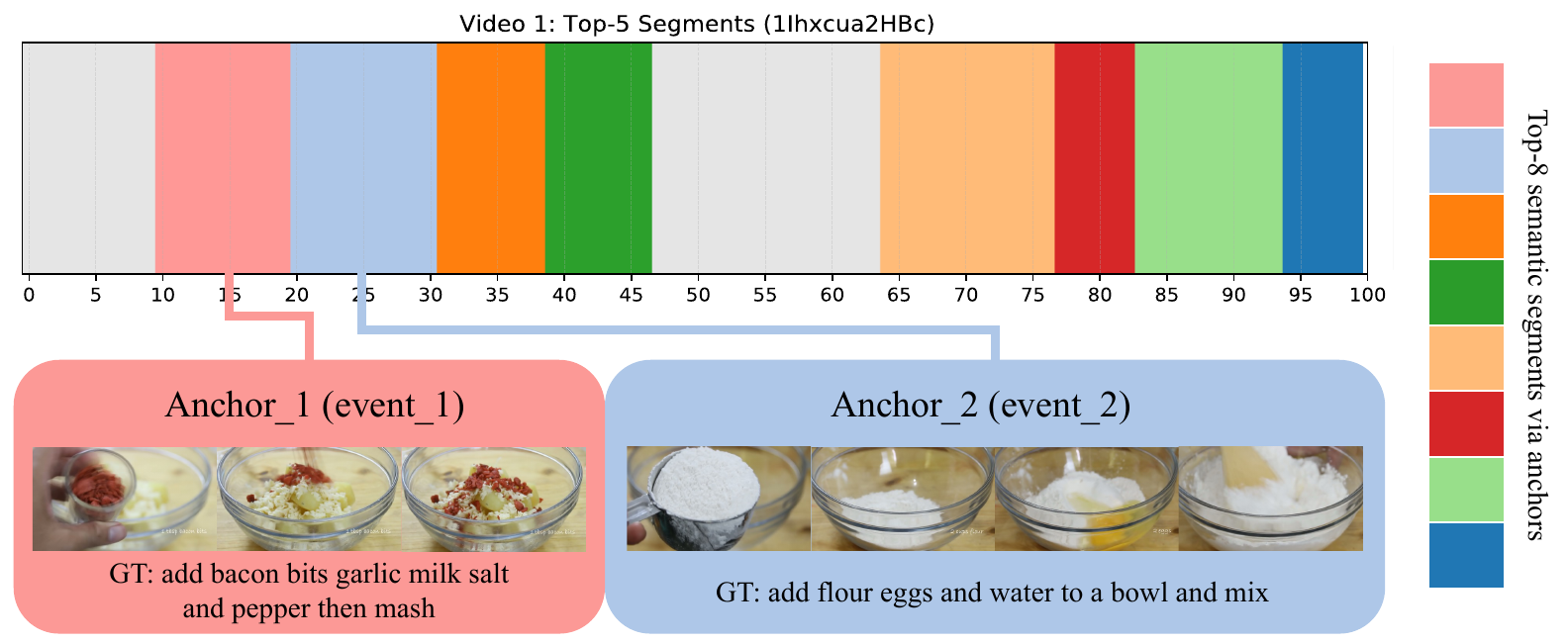}
\caption{\textbf{Visualization of the Semantic Prototypes.}}
\label{supp:semantic_proto}
\end{figure}

\paragraph{Segment and Retrieved Captions Qualitative Results.}
\label{supp:ot_qualitative}
We qualitatively compare segments and retrieved captions in~\Cref{supp:seg_qual}. While prior methods~\cite{kim2025hicm2, jeon2025sali4vid} often produce misaligned or drifting segment boundaries, our saliency-guided transport yields segments that align more closely with ground truth event durations. The retrieved captions are also semantically more relevant to the video content, validating the effectiveness of \textit{SGSR}. In \textit{SGSR}, we define $K$ learnable anchor embeddings as semantic prototypes, where each anchor serves as a general temporal placeholder representing an individual event. During optimal transport, frames are assigned to the anchor with the highest feature similarity, forming temporally coherent segments. We visualize the learned anchor assignments in~\Cref{supp:semantic_proto}, showing that the anchors consistently capture meaningful event boundaries.
\begin{figure}[t]
    \centering
    \includegraphics[width=0.98\linewidth]{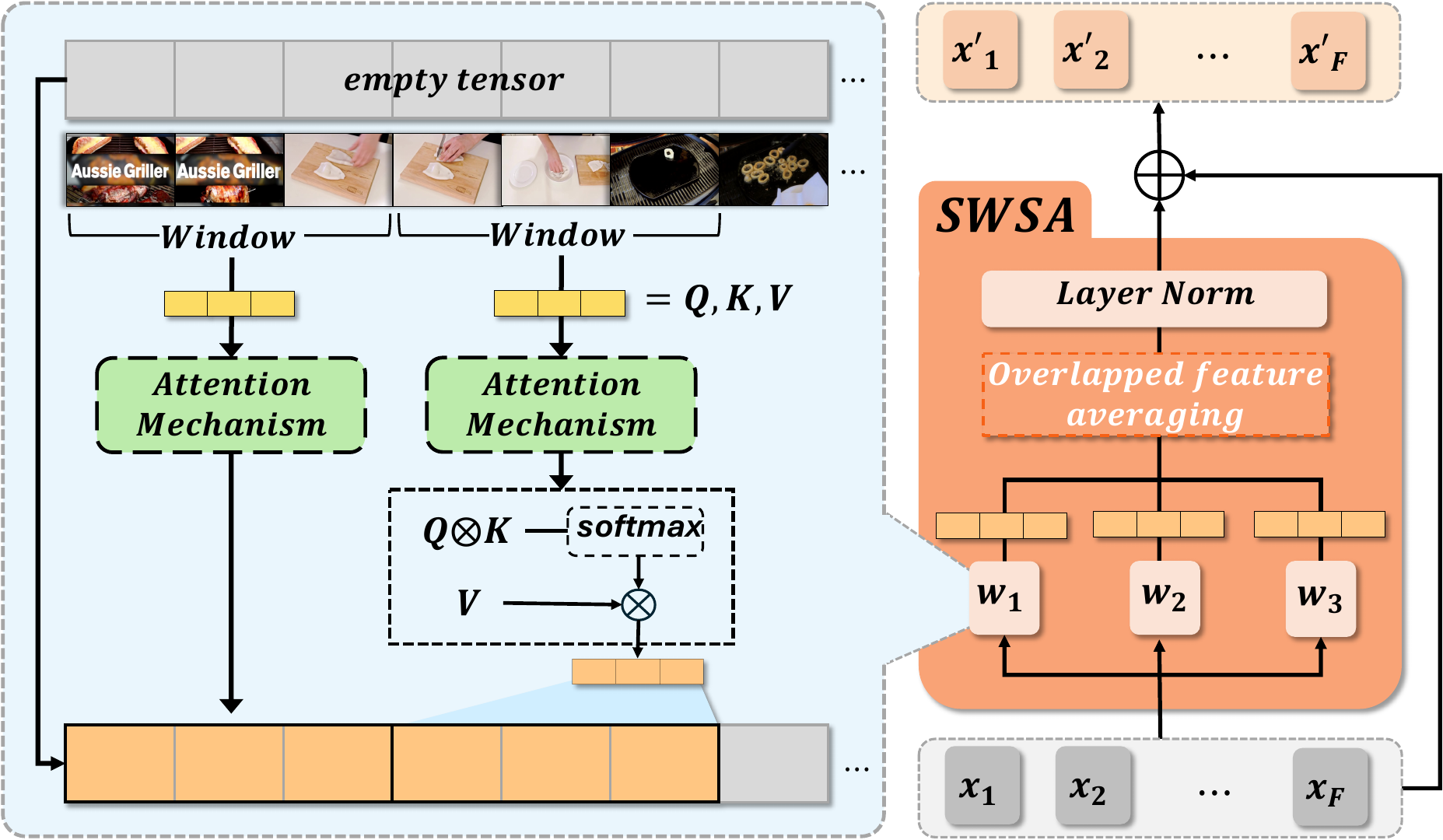}
    \caption{
    \textbf{Architecture of the \textit{SWSA} module.}
    \textit{SWSA} refines $X$ through local self-attention without linear projections over sliding windows, where overlapping regions are averaged and residually added to produce refined features $X'$.}
\label{fig:SWSA}
\end{figure}

\begin{figure*}[t]
\begin{center}
\includegraphics[width=0.98\textwidth]{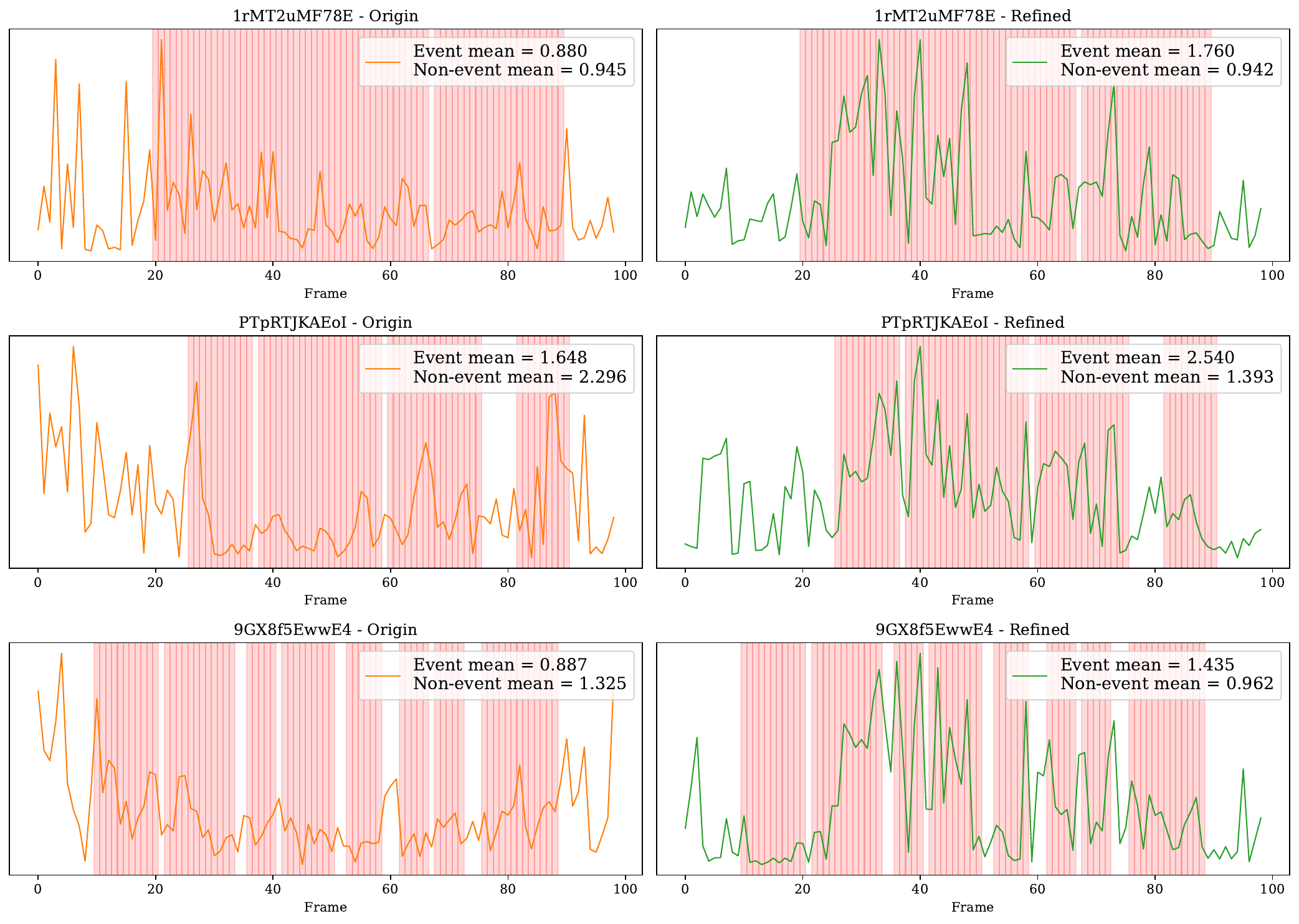}
\end{center}
\vspace{-3mm}
\caption{\textbf{Frame transition sharpness comparison between original and refined features}. 
For each video in the YouCook2 validation set, red spans indicate ground truth event regions. 
Refined features exhibit larger transition magnitudes within event intervals, indicating stronger semantic contrast and clearer event boundaries than original features.}
\label{supp:x_x'_diff}
\end{figure*}

\section{Additional Analysis and Experiments}
\label{supp:add_exp}

\paragraph{Feature Refinement.}
The architecture of \textit{SWSA} is on the ~\Cref{fig:SWSA}.

Given the input features $X \in \mathbb{R}^{F \times D}$ and a set of window sizes $\{w_1, w_2, w_3\}$, we compute local self-attention within each window. For a window of size $w$, the segment starting at position $i$ is denoted as $X_{i:i+w} \in \mathbb{R}^{w \times D}$. The attention output for this segment is computed as:
\begin{equation}
    A_{i:i+w} = \text{softmax}\left(\frac{X_{i:i+w} X_{i:i+w}^\top}{\sqrt{D}}\right) X_{i:i+w}.
\end{equation}

Since windows overlap, each frame $n$ may appear in multiple windows. We average all attention outputs that cover frame $n$, weighted by the number of overlapping windows $C_n$:
\begin{equation}
    \hat{X}_n = \frac{1}{C_n} \sum_{(w,i): \, n \in [i, i+w)} A_{n}^{(w,i)},
\end{equation}
where $C_n$ is the number of windows covering frame $n$, and $A_{n}^{(w,i)}$ is the attention output at position $n$ from the window of size $w$ starting at $i$. The final refined representation is obtained via a residual connection:
\begin{equation}
    X' = X + \text{LayerNorm}(\hat{X}).
\end{equation}

\paragraph{Feature Difference between Training and Inference.}
In \textit{STaRC}, we use refined features from \textit{SWSA} differently during training and inference. During training, the refined features $X'$ are used only as input to the highlight detection module, while the decoder receives original features $X$. During inference, the refined features $X'$ are used both as input to the highlight detection module and as decoder input. 
To analyze the refinement impact, we measure frame-to-frame transition sharpness in~\Cref{supp:x_x'_diff}, by computing L2 distances between consecutive frame embeddings. 
Refined features $X'$ produce sharper transitions at event boundaries compared to $X$, while non-event regions show smaller transitions.

\paragraph{Comparison with Vision-Language Models.}
\begin{table}[t] 
\centering
\resizebox{\columnwidth}{!}{ 
\scriptsize
\begin{tabular}{l|c|ccc}
\toprule[1.5pt]
\textbf{Method} & \textbf{PT} & \textbf{CIDEr} & \textbf{SODA\_c} & \textbf{F1} \\
\midrule
TimeChat~\pub{CVPR24}   & \ding{51} & 11.0  & 3.4  & 19.5 \\
VTG-LLM~\pub{AAAI25}    & \ding{51} & 13.4  & 3.6  & 20.6 \\
TRACE~\pub{ICLR25}      & \ding{51} & 35.5  & 6.7  & 31.8 \\
TimeExpert~\pub{ICCV25} & \ding{51} & 39.0  & 7.2  & 33.5 \\
\midrule
\rowcolor{gray!10} 
\textbf{Ours}           & \ding{51} & \textbf{80.53} & \textbf{10.73} & \textbf{34.34} \\
\bottomrule[1.5pt]
\end{tabular}}
\caption{Comparison with VLM methods on the YouCook2 validation set}
\label{tab:VLM_model}
\end{table}
We also compare with recent Vision-Language Models (VLMs) on the DVC task. As shown in~\Cref{tab:VLM_model}, even powerful VLMs struggle to produce accurate event boundaries and captions simultaneously. This suggests that general-purpose VLMs are not yet sufficient for DVC, and task-specific architectures remain essential.
\section{Failure case and limitations}
\label{supp:limitation}

\begin{figure*}[t]
\begin{center}
\includegraphics[width=0.98\textwidth]{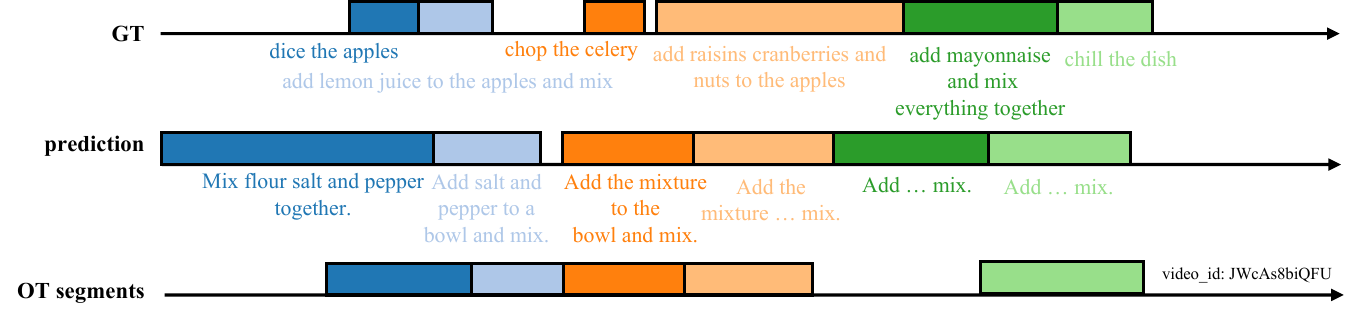}
\end{center}
\vspace{-3mm}
\caption{\textbf{Failure case of \textit{STaRC}}.}
\label{supp:fail}
\end{figure*}

\textit{STaRC} occasionally produces redundant captions despite precise localization (a challenge shared by existing DVC methods). OT-based retrieval is effective overall but can introduce noise in some segments (\Cref{supp:fail}), which we leave for future work.

\begin{figure*}[t]
\centering

\begin{subfigure}{0.98\textwidth}
    \centering
    \includegraphics[width=\textwidth]{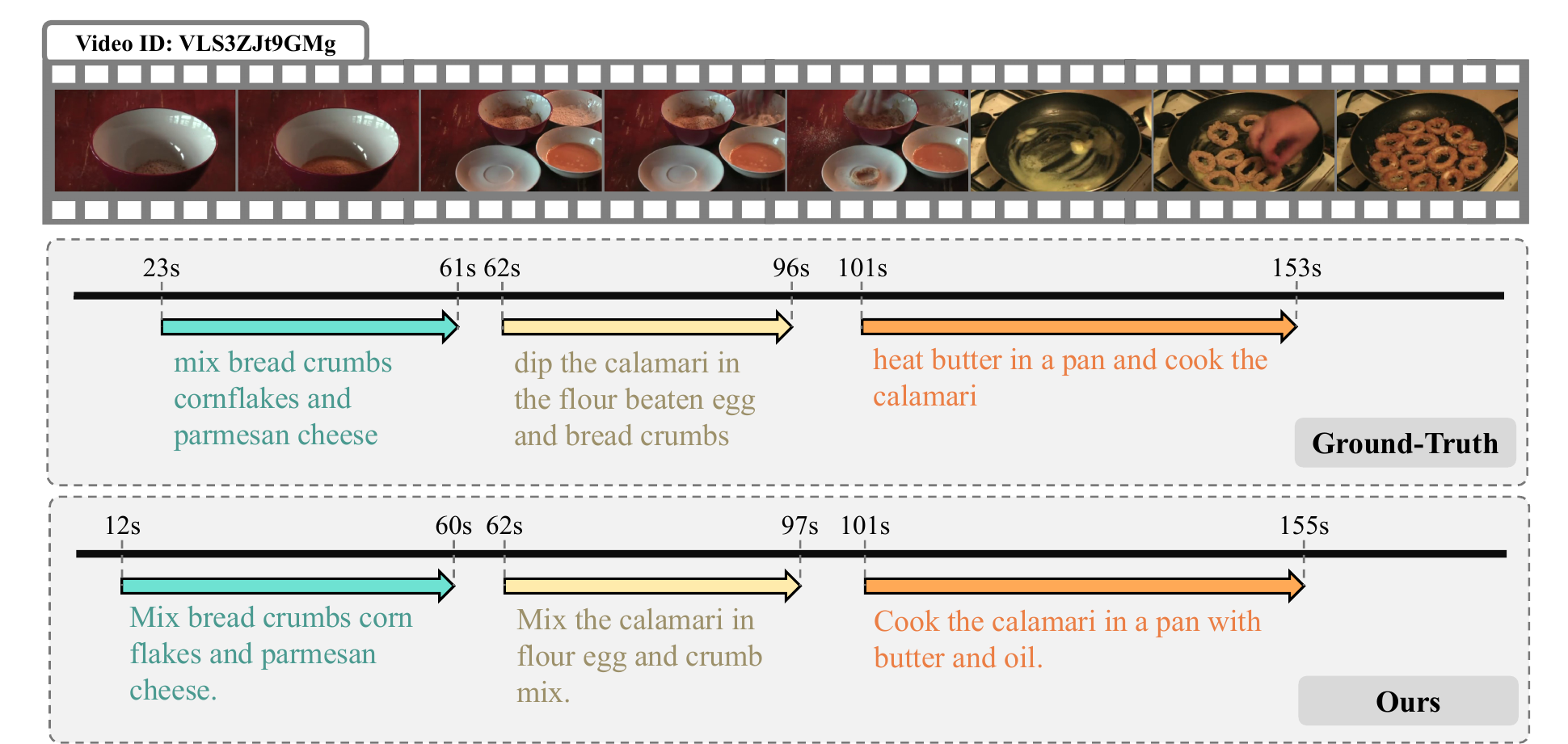}
    \label{supp:cap_qual_1_sub}
\end{subfigure}

\vspace{4mm}

\begin{subfigure}{0.98\textwidth}
    \centering
    \includegraphics[width=\textwidth]{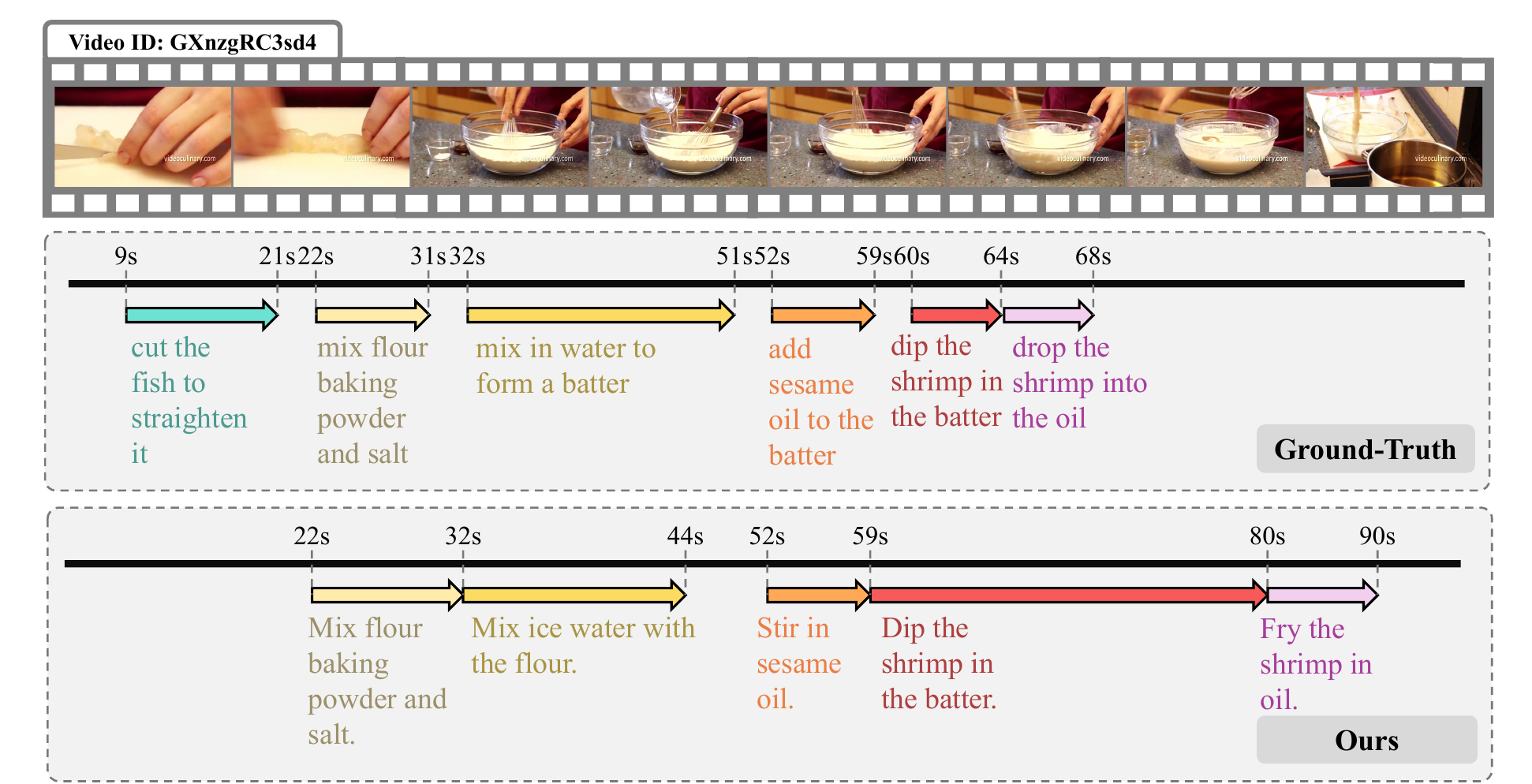}
    \label{supp:cap_qual_2_sub}
\end{subfigure}

\caption{
{\textbf{Additional qualitative results for \textit{STaRC} on the YouCook2 validation set.}}
}
\label{supp:qual_combined}
\end{figure*}

\end{document}